\documentclass[10pt,journal,compsoc]{IEEEtran}

\ifCLASSOPTIONcompsoc
  \usepackage[nocompress]{cite}
\else
  \usepackage{cite}
\fi

\ifCLASSINFOpdf
  \usepackage[pdftex]{graphicx}
\else
  \usepackage[dvips]{graphicx}
\fi

\ifCLASSOPTIONcompsoc
  \usepackage[caption=false,font=footnotesize,labelfont=sf,textfont=sf]{subfig}
\else
  \usepackage[caption=false,font=footnotesize]{subfig}
  \usepackage{textcomp}
\fi

\usepackage{tikz, array, multirow, comment}
\usetikzlibrary{trees}
\usepackage[font=small,skip=10pt]{caption}

\begin{document}
\title{Survey on RGB, 3D, Thermal, and Multimodal Approaches for Facial Expression Recognition: History, Trends, and Affect-related Applications}

\author{Ciprian~A.~Corneanu,~Marc~Oliu,~Jeffrey~F.~Cohn,~and~Sergio~Escalera%
\IEEEcompsocitemizethanks{\IEEEcompsocthanksitem C. Corneanu, M. Oliu and S. Escalera are with the Computer Vision Center, UAB, Barcelona, Spain, and with the Dept. Applied Methematics, University of Barcelona, Spain.\protect\\
E-mail: cipriancorneanu@ub.edu, moliusimon@gmail.com, sergio@maia.ub.es, 
\IEEEcompsocthanksitem J. F. Cohn is with the Robotics Institute, CMU, Pittsburgh, Pennsylvania, and with the Dept. Psychology, University of Pittsburgh, Pennsylvania.\protect\\
E-mail: jeffcohn@cs.cmu.edu} 
\thanks{Manuscript received March 8, 2015; revised XX, 2015.}}


\IEEEtitleabstractindextext{%
\begin{abstract}
Facial expressions are an important way through which humans interact socially. Building a system capable of automatically recognizing facial expressions from images and video has been an intense field of study in recent years. Interpreting such expressions remains challenging and much research is needed about the way they relate to human affect. This paper presents a general overview of automatic RGB, 3D, thermal and multimodal facial expression analysis. We define a new taxonomy for the field, encompassing all steps from face detection to facial expression recognition, and describe and classify the state of the art methods accordingly. We also present the important datasets and the bench-marking of most influential methods.  We conclude with a general discussion about trends, important questions and future lines of research.
\end{abstract}

}

\maketitle
\IEEEdisplaynontitleabstractindextext
\IEEEpeerreviewmaketitle


\IEEEraisesectionheading{\section{Introduction}\label{sec:introduction}}

\IEEEPARstart{F}{acial} expressions (FE) are vital signaling systems of affect, conveying cues about the emotional state of persons. Together with voice, language, hands and posture of the body, they form a fundamental communication system between humans in social contexts. Automatic FE recognition (AFER) is an interdisciplinary domain standing at the crossing of behavioral science, neurology, and artificial intelligence.

\begin{figure} 
\centering
\includegraphics[width=\linewidth]{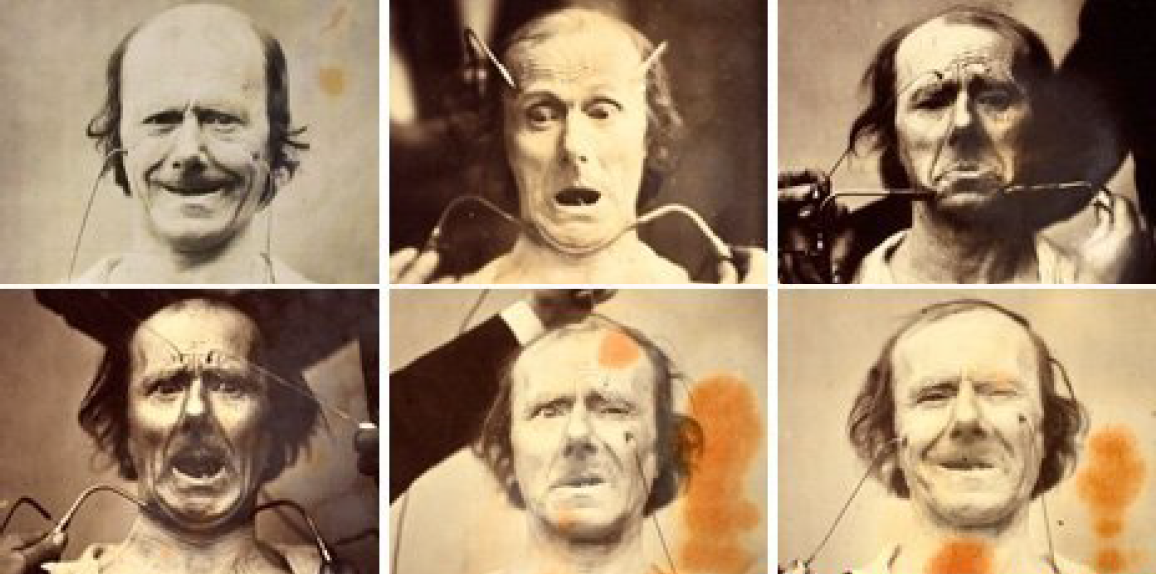}
\vspace{-0.7cm}
\caption{\scriptsize{In the 19th century, Duchenne de Boulogne conducted experiments on how FEs are produced. From \cite{duchenne1990mechanism}.}}
\vspace{-0.5cm}
\label{fig:duchenne}
\end{figure}

Studies of the face were greatly influenced in premodern times by popular theories of physiognomy and creationism. Physiognomy assumed that a person's character or personality could be judged by their outer appearance, especially the face \cite{physiognomy}. Leonardo Da Vinci was one of the first to refute such claims stating they were without scientific support \cite{chastel02}. In the 17th century in England, John Buwler studied human communication with particular interest in the sign language of persons with hearing impairment. His book \emph{Pathomyotomia} or \emph{Dissection of the significant Muscles of the Affections of the Mind} was the first consistent work in the English language on the muscular mechanism of FE \cite{greenblatt94}. About two centuries later, influenced by creationism, Sir Charles Bell investigated FE as part of his research on sensory and motor control. He believed that FE was endowed by the Creator solely for human communication. Subsequently, Duchenne de Boulogne conducted systematic studies on how FEs are produced \cite{duchenne1990mechanism}. He published beautiful pictures of sometimes strange FEs obtained by electrically stimulating facial muscles (see Figure \ref{fig:duchenne}). Approximately in the same historical period, Charles Darwin firmly placed FE in an evolutionary context \cite{darwin1872}. This marked the beginning of modern research of FEs. More recently, important advancements were made through the works of researchers like Carroll Izard and Paul Ekman who inspired by Darwin performed seminal studies of FEs \cite{izard71, ekman71, ekman79}.

In the last years excellent surveys on automatic facial expression analysis have been published \cite{zeng09, salah11, sandbach12sota, sariyanidi2014automatic}. For a more processing oriented review of the literature the reader is mainly referred to \cite{sariyanidi2014automatic, salah11}. For an introduction into AFER in natural conditions the reader is referred to \cite{zeng09}. Readers interested mainly in 3D AFER, should refer to the work of Sandbach et al. \cite{sandbach12sota}.

In this survey, we define a comprehensive taxonomy of automatic RGB\footnote{RGB: Additive color model in which red, green, and blue light are combined to reproduce a broad array of colors.}, 3D, thermal, and multimodal computer vision approaches for AFER. The definition and choices of the different components are analyzed and discussed. This is complemented with a section dedicated to the historical evolution of FE approaches and an in-depth analysis of latest trends. Additionally, we provide an introduction into affect inference from the face from a evolutionary perspective. We emphasize research produced since the last major review of AFER in 2009 \cite{zeng09}. Our focus on inferring affect, defining a comprehensive taxonomy and treating different modalities is aiming at proposing a more general perspective on AFER and its current trends. 

The paper is organized as follows: Section \ref{sec:inference} discusses affect in terms of FEs. Section \ref{sec:taxonomy} presents a taxonomy of automatic RGB, 3D, thermal and multimodal recognition of FEs.  Section \ref{sec:methods} reviews the historical evolution in AFER and focuses on recent important trends. Finally, Section \ref{sec:discussion} concludes with a general discussion. 

\section{Inferring affect from FEs}
\label{sec:inference}

Depending on context FEs may have varied communicative functions. They can regulate conversations by signaling turn-taking, convey biometric information, express intensity of mental effort, and signal emotion. By far, the latter has been the one most studied.

\vspace{-0.2cm}\subsection{Describing affect}
\label{sec:inference:theories}

Attempts to describe human emotion mainly fall into two approaches: categorical and dimensional description.

\begin{figure}[ht!] 
\centering
\includegraphics[width=\linewidth]{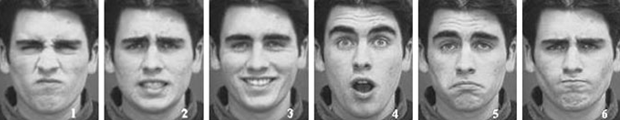}
\vspace{-0.7cm}\caption{\scriptsize{Primary emotions expressed on the face. From left to right: disgust, fear, joy, surprise, sadness, anger. From \cite{what-when-how}.}}
\label{fig:basic_emotions}
\vspace{-0.2cm}\end{figure}

\textbf{Categorical description of affect.} Classifying emotions into a set of distinct classes that can be recognized and described easily in daily language has been common since at least the time of Darwin. More recently, influenced by the research of Paul Ekman \cite{ekman71, ekman94} a dominant view upon affect is based on the underlying assumption that humans universally express a set of discrete primary emotions which include happiness, sadness, fear, anger, disgust, and surprise (see Figure \ref{fig:basic_emotions}). Mainly because of its simplicity and its universality claim, the universal primary emotions hypothesis has been extensively exploited in affective computing.

\textbf{Dimensional description of affect.} Another popular approach is to place a particular emotion into a space having a limited set of dimensions \cite{greenwald89, russell77, watson88}. These dimensions include valence (how pleasent or unpleasent a feeling is) activation\footnote{Also known as arousal.} (how likely is the person to take action under the emotional state) and control (the sense of control over the emotion). Due to the higher dimensionality of such descriptions they can potentially describe more complex and subtle emotions. Unfortunately, the richness of the space is more difficult to use for automatic recognition systems because it can be challenging to link such described emotion to a FE. Usually automatic systems based on dimensional representation of emotion simplify the problem by dividing the space in a limited set of categories like positive vs negative or quadrants of the 2D space \cite{zeng09}.

\vspace{-0.2cm}\subsection{An evolutionist approach to FE of affect}
\label{sec:inference:evolution}

At the end of the 19th century Charles Darwin wrote \emph{The Expression of the emotion in Man and Animals}, which largely inspired the study of FE of emotion. Darwin proposed that FEs are the residual actions of more complete behavioral responses to environmental challenges. Constricting the nostrils in disgust served to reduce inhalation of noxious or harmful substances. Widening the eyes in surprise increased the visual field to see an unexpected stimulus. Darwin emphasized the adaptive functions of FEs.

More recent evolutionary models have come to emphasize their communicative functions \cite{fridlund1992}. \cite{shariff2011} proposed a process of exaptation in which adaptations (such as constricting the nostrils in disgust) became recruited to serve communicative functions. Expressions (or displays) were ritualized to communicate information vital to survival. In this way, two abilities were selected for their survival advantages. One was to automatically display exaggerated forms of the original expressions; the other was to automatically interpret the meaning of these expressions. From this perspective, disgust communicates potentially aversive foods or moral violations; sadness communicates request for comfort. While some aspects of evolutionary accounts of FE are controversial \cite{barrett2011}, strong evidence exists in their support. Evidence includes universality of FEs of emotion, physiological specificity of emotion, and automatic appraisal and unbidden occurrence \cite{ekman1992,ekman2000,matsumoto08}.

\emph{Universality}. There is a high degree of consistency in the facial musculature among peoples of the world. The muscles necessary to express primary emotions are found universally \cite{schmidt2001,gray66,burrows2014}, and homologous muscles have been documented in non-human primates \cite{waller2008,waller2012,waller2008mapping}. Similar FEs in response to species-typical signals have been observed in both human and non-human primates \cite{eibl1989}.

\emph{Recognition}. Numerous perceptual judgment studies support the hypothesis that FEs are interpreted similarly at levels well above chance in both Western and non-Western societies. Even critics of strong evolutionary accounts \cite{russell1994}, \cite{jack2009} find that recognition of FEs of emotion are universally above chance and in many cases quite higher.  

\emph{Physiological specificity}. Physiological specificity appears to exist as well. Using directed facial action tasks to elicit basic emotions, Levenson and colleagues \cite{levenson1990} found that HR, GSR, and skin temperature systematically varied with the hypothesized functions of basic emotions. In anger, blood flow to the hands increased to prepare for fight. For the central nervous system, patterns of prefrontal and temporal asymmetry systematically differed between enjoyment and disgust when measured using the \emph{Facial Action Coding System} (FACS) \cite{ekman1990}. Left-frontal asymmetry was greater during enjoyment; right frontal asymmetry was greater during disgust. These findings support the view that emotion expressions reliably signal action tendencies \cite{frijda1997,niedenthal2007}.

\emph{Subjective experience}. While not critical to an evolutionary account of emotion, evidence exists as well for concordance between subjective experience and FE of emotion \cite{ekmanRosenberg2005, bp4d}.  However, more work is needed in this regard. Until recently, manual annotation of FE or facial EMG were the only means to measure FE of emotion. Because manual annotation is labor intensive, replication of studies is limited.

In summary, the study of FE initially was strongly motivated by evolutionary accounts of emotion. Evidence has broadly supported those accounts. However, FE more broadly figures in cultural bio-psycho-social accounts of emotion. Facial expression signals emotion, communicative intent, individual differences in personality, and psychiatric and medical status, and helps to regulate social interaction. With the advent of automated methods of AFER, we are poised to make major discoveries in these areas. 

\vspace{-0.2cm}\subsection{Applications}
\label{sec:inference:applications}

The ability to automatically recognize FEs and infer affect has a wide range of applications. AFER, usually combined with speech, gaze and standard interactions like mouse movements and keystrokes can be used to build adaptive environments by detecting the user's affective states \cite{duric2002integrating, maat2007gaze}. Similarly, one can build socially aware systems \cite{vinciarelli09, devault2014}, or robots with social skills like Sony's AIBO and ATR's Robovie \cite{ishiguro2001robovie}. Detecting students' frustration can help improve e-learning experiences \cite{kapoor2007automatic}. Gaming experience can also be improved by adapting difficulty, music, characters or mission according to the player's emotional responses \cite{bakkes2012personalised, tan2012feasibility,blom2014towards}. Pain detection is used for monitoring patient progress in clinical settings \cite{lucey10,kaltwang2012continous, iranispatiotemporal}. Detection of truthfulness or potential deception can be used during police interrogations or job interviews \cite{ryan09}. Monitoring drowsiness or attentive and emotional status of the driver is critical for the safety and comfort of driving \cite{vural2007drowsy}. Depression recognition from FEs is a very important application in analysis of psychological distress \cite{girard2014nonverbal, scherer2013automatic, joshi2012neural}. Finally, in recent years successful commercial applications like Emotient \cite{emotient}, Affectiva \cite{affectiva}, RealEyes \cite{realeyes} and Kairos \cite{kairos} perform large-scale internet-based assessments of viewer reactions to ads and related material for predicting buying behaviour. 

\vspace{-0.3cm}
\section{A taxonomy for recognizing FEs} 
\label{sec:taxonomy}


\definecolor{tred}{rgb}{0.6,0,0}    \newcommand{\crgb}[1]{\textcolor{tred}{#1}}
\definecolor{tgreen}{rgb}{0,0.4,0}  \newcommand{\cdepth}[1]{\textcolor{tgreen}{#1}}
\definecolor{torange}{rgb}{0.4,0,0.4} \newcommand{\cthermal}[1]{\textcolor{torange}{#1}}
\definecolor{tblue}{rgb}{0,0,0.6} \newcommand{\cnocat}[1]{\textcolor{tblue}{#1}}
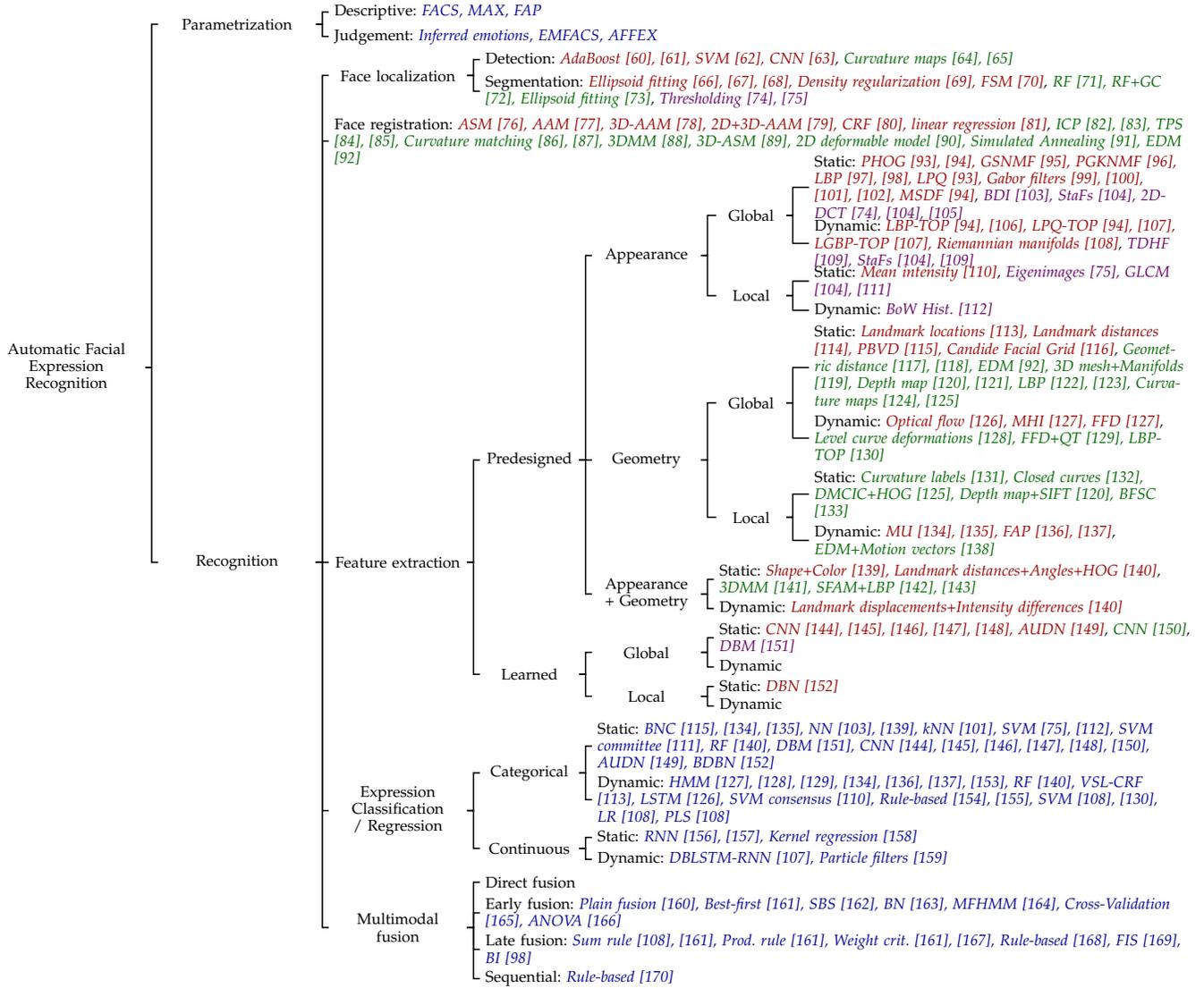
\begin{figure*}
\centering
\resizebox{\textwidth}{!}{%
\begin{tikzpicture}[grow'=right,level distance=1.65in]
	\tikzstyle{edge from parent} = [thick, draw, edge from parent fork right]
	\tikzstyle{every node}=[font=\small]
	\tikzstyle{root} = [text width=30mm,align=center]
	\tikzstyle{nodeL1} = [text width=30mm,align=center] \tikzstyle{level 1} = [level distance=35mm]
	\tikzstyle{nodeL2} = [text width=26mm,align=center,xshift=-2mm] \tikzstyle{level 2} = [level distance=35mm]
	\tikzstyle{nodeL3} = [text width=18mm,align=center,xshift=-4mm] \tikzstyle{level 3} = [level distance=31mm]
	\tikzstyle{nodeL4} = [text width=20mm,align=center,xshift=1mm] \tikzstyle{level 4} = [level distance=23mm]
	\tikzstyle{nodeL5} = [text width=12mm,align=center,xshift=-3.5mm] \tikzstyle{level 5} = [level distance=25mm]
	\tikzstyle{nodeL6} = [text width=12mm,align=center,xshift=-3mm] \tikzstyle{level 6} = [level distance=17mm]
	\tikzstyle{termL2} = [text width=175mm,align=left,xshift=72.5mm]
	\tikzstyle{termL3} = [text width=145mm,align=left,xshift=59.5mm]
	\tikzstyle{termL4} = [text width=120mm,align=left,xshift=51mm]
	\tikzstyle{termL5} = [text width=100mm,align=left,xshift=40mm]
	\tikzstyle{termL6} = [text width=76mm,align=left,xshift=34mm]
    
    \node[root]{Automatic Facial Expression Recognition}
    child[sibling distance=81mm]{
    	node[nodeL1,yshift=30mm]{Parametrization}
    	child[sibling distance=5mm]{
    	    node[termL2]{Descriptive:  \cnocat{\emph{FACS, MAX, FAP}}}
    	}child[sibling distance=5mm]{
    	    node[termL2]{Judgement: \cnocat{\emph{Inferred emotions, EMFACS, AFFEX}}}
    	}
    }child[sibling distance=80mm]{
    	node[nodeL1]{Recognition}
    	child[sibling distance=50mm]{
    		node[nodeL2]{Face localization}
   			child[sibling distance=6.5mm]{
   				node[termL3]{Detection: \emph{\crgb{AdaBoost \cite{viola2001rapid,jones2003fast}, SVM \cite{dalal2005histograms}, CNN \cite{osadchy2007synergistic}}, \cdepth{Curvature maps \cite{colombo20063d,nair20093}}}}
   			}child[sibling distance=6.5mm]{
   				node[termL3]{Segmentation: \emph{\crgb{Ellipsoid fitting \cite{sobottka1996segmentation,sirohey1998human,sobottka1998novel}, Density regularization \cite{chai1999face}, FSM \cite{li2008saliency}}, \cdepth{RF \cite{shotton2013real}, RF+GC \cite{hernandez2012graph}, Ellipsoid fitting \cite{pamplona2010automatic}}, \cthermal{Thresholding \cite{koda2009facial,trujillo2005automatic}}}}
   			}
    	}child[sibling distance=86.5mm]{
    		node[termL2]{Face registration: \emph{\crgb{ASM \cite{cootes1995active}, AAM \cite{cootes2001active}, 3D-AAM \cite{romdhani2003efficient}, 2D+3D-AAM \cite{baker2004lucas}, CRF \cite{dantone2012real}, linear regression \cite{xiong2013supervised}}, \cdepth{ICP \cite{besl1992method,alyuz2012adaptive}, TPS \cite{mao2004constructing,tena2006validated}, Curvature matching \cite{szeptycki2009coarse,alyuz2010regional}, 3DMM \cite{blanz1999morphable}, 3D-ASM \cite{fanelli2013real}, 2D deformable model \cite{savran2008non}, Simulated Annealing \cite{queirolo20103d}, EDM \cite{mpiperis20083d}}}}
    	}child[]{
    		node[nodeL2]{Feature extraction}
    		child[sibling distance=42mm]{
    			node[nodeL3]{Predesigned}
				child[sibling distance=42mm]{
	   				node[nodeL4]{Appearance}
	   				child[sibling distance=16.5mm]{
	   					node[nodeL5]{Global}
	   					child[sibling distance=11.5mm]{
	   						node[termL6]{Static: \emph{\crgb{PHOG \cite{dhall2011emotion,sun2014combining}, GSNMF \cite{zhi2011graph}, PGKNMF \cite{zafeiriou2009nonlinear}, LBP \cite{shan09,savran2014temporal}, LPQ \cite{dhall2011emotion}, Gabor filters \cite{littlewort04,littlewort2011computer,gu2012facial,lyons99}, MSDF \cite{sun2014combining}}, \cthermal{BDI \cite{yoshitomi1997facial}, StaFs \cite{wang2014emotion}, 2D-DCT \cite{koda2009facial,yoshitomi2010facial,wang2014emotion}}}}
	   					}child[sibling distance=11.5mm]{
	   						node[termL6]{Dynamic: \emph{\crgb{LBP-TOP \cite{zhao2007dynamic,sun2014combining}, LPQ-TOP \cite{sun2014combining, he2015multimodal}, LGBP-TOP \cite{he2015multimodal}, Riemannian manifolds \cite{liu2014combining}}, \cthermal{TDHF \cite{liu2011emotion}, StaFs \cite{liu2011emotion,wang2014emotion}}}}
	   					}
	   				}child[sibling distance=16.5mm]{
	   					node[nodeL5]{Local}
						child[sibling distance=6mm]{
	   						node[termL6]{Static: \emph{\crgb{Mean intensity \cite{geetha2009facial}}, \cthermal{Eigenimages \cite{trujillo2005automatic}, GLCM \cite{hernandez2007visual,wang2014emotion}}}}
	   					}child[sibling distance=6mm]{
	   						node[termL6]{Dynamic: \emph{\cthermal{BoW Hist. \cite{liu2015spontaneous}}}}
	   					}
	   				}
	   			}child[sibling distance=25mm]{
	   				node[nodeL4]{Geometry}
	   				child[sibling distance=23.5mm]{
	   					node[nodeL5]{Global}
	   					child[sibling distance=14.5mm]{
	   						node[termL6]{Static: \emph{\crgb{Landmark locations \cite{walecki15}, Landmark distances \cite{pantic06}, PBVD \cite{sebe07}, Candide Facial Grid \cite{kotsia07}}, \cdepth{Geometric distance \cite{tang2008segments,tang2008automatic}, EDM \cite{mpiperis20083d}, 3D mesh+Manifolds \cite{chang05}, Depth map \cite{berretti20113d,vretos11}, LBP \cite{sandbach12lbp,hayat13}, Curvature maps \cite{zeng13,lemaire2013fully}}}}
	   					}child[sibling distance=14.5mm]{
	   						node[termL6]{Dynamic: \emph{\crgb{Optical flow \cite{wollmer2013lstm}, MHI \cite{koelstra10}, FFD \cite{koelstra10}}, \cdepth{Level curve deformations \cite{le11}, FFD+QT \cite{sandbach11}, LBP-TOP \cite{fang12}}}}
	   					}
	   				}child[sibling distance=23.5mm]{
                                                               node[nodeL5]{Local}
						child[sibling distance=9.5mm]{
	   						node[termL6]{Static: \emph{\cdepth{Curvature labels \cite{wang20063d}, Closed curves \cite{maalej2011shape}, DMCIC+HOG \cite{lemaire2013fully}, Depth map+SIFT \cite{berretti20113d}, BFSC \cite{gong2009automatic}}}}
	   					}child[sibling distance=9.5mm]{
	   						node[termL6]{Dynamic:  \emph{\crgb{MU \cite{cohen03,cohen03learning}, FAP \cite{pardas02,aleksic06}}, \cdepth{EDM+Motion vectors \cite{yin2006analyzing}}}}
	   					}
	   				}
	   			}child[sibling distance=27.5mm]{
	   				node[nodeL4]{Appearance + Geometry}
	   				child[sibling distance=6mm]{
	   					node[termL5]{Static: \emph{\crgb{Shape+Color \cite{tian01}, Landmark distances+Angles+HOG \cite{dapogny2015dynamic}}, \cdepth{3DMM \cite{ramanathan2006human}, SFAM+LBP \cite{zhao2010automatic,zhao2013unified}}}}
	   				}child[sibling distance=6mm]{
	   					node[termL5]{Dynamic: \emph{\crgb{Landmark displacements+Intensity differences \cite{dapogny2015dynamic}}}}
	   				}
	   			}
    		}child[sibling distance=46mm]{
    			node[nodeL3]{Learned}
    			child[sibling distance=9mm]{
    				node[nodeL4]{Global}
    				child[sibling distance=6mm]{
    					node[termL5]{Static: \emph{\crgb{CNN \cite{ranzato11,rifai12,liu2014learning,kahou2013combining,song2014deep}, AUDN \cite{liu2013aware}}, \cdepth{CNN \cite{ijjina2014facial}}, \cthermal{DBM \cite{he2013facial}}}}
    				}child[sibling distance=6mm]{
    					node[termL5]{Dynamic}
    				}
    			}child[sibling distance=9mm]{
    				node[nodeL4]{Local}
    				child[sibling distance=4mm]{
    					node[termL5]{Static: \emph{\crgb{DBN \cite{liu2014facial}}}}
    				}child[sibling distance=4mm]{
    					node[termL5]{Dynamic}
    				}
    			}
    		}
    	}child[sibling distance=51mm]{
    		node[nodeL2]{Expression Classification / Regression}
    		child[sibling distance=15.5mm]{
    			node[nodeL3]{Categorical}
    			child[sibling distance=11mm]{
    				node[termL4]{Static: \cnocat{\emph{BNC \cite{cohen03,cohen03learning,sebe07}, NN \cite{yoshitomi1997facial,tian01}, kNN \cite{gu2012facial}, SVM \cite{trujillo2005automatic,liu2015spontaneous}, SVM committee \cite{hernandez2007visual}, RF \cite{dapogny2015dynamic}, DBM \cite{he2013facial}, CNN \cite{ranzato11,rifai12,liu2014learning,kahou2013combining,ijjina2014facial,song2014deep}, AUDN \cite{liu2013aware}, BDBN \cite{liu2014facial}}}}
    			}child[sibling distance=11mm]{
    				node[termL4]{Dynamic: \cnocat{\emph{HMM \cite{cohen03,aleksic06,koelstra10,le11,sandbach11,wu2015multi,pardas02}, RF \cite{dapogny2015dynamic}, VSL-CRF \cite{walecki15}, LSTM \cite{wollmer2013lstm}, SVM consensus \cite{geetha2009facial}, Rule-based \cite{tsalakanidou2010real,tsalakanidou2009robust}, SVM \cite{liu2014combining,fang12}, LR \cite{liu2014combining}, PLS \cite{liu2014combining}}}}
    			}
    		}child[sibling distance=15.5mm]{
    			node[nodeL3]{Continuous}
    			child[sibling distance=4.5mm]{
    				node[termL4]{Static: \cnocat{\emph{RNN \cite{fragopanagos2005emotion,caridakis2006modeling}, Kernel regression \cite{nicolle2012robust}}}}
    			}child[sibling distance=4.5mm]{
    				node[termL4]{Dynamic: \cnocat{\emph{DBLSTM-RNN \cite{he2015multimodal}, Particle filters \cite{savran2012combining}}}}
    			}
    		}
    	}child[sibling distance=37.5mm]{
    		node[nodeL2]{Multimodal fusion}
    		child[sibling distance=6.2mm]{
    			node[termL3]{Direct fusion}
    		}child[sibling distance=6mm]{
    			node[termL3]{Early fusion: \cnocat{\emph{Plain fusion \cite{huang1998bimodal}, Best-first \cite{gunes2005affect}, SBS \cite{busso2004analysis}, BN \cite{sebe2006emotion}, MFHMM \cite{zeng2008audio}, Cross-Validation \cite{kessous2010multimodal}, ANOVA \cite{d2010multimodal}}}}
    		}child[sibling distance=9mm]{
    			node[termL3]{Late fusion: \cnocat{\emph{Sum rule \cite{gunes2005affect,liu2014combining}, Prod. rule \cite{gunes2005affect}, Weight crit. \cite{gunes2005affect,yoshitomi2000effect}, Rule-based \cite{de2000bimodal}, FIS \cite{soladie2012multimodal}, BI \cite{savran2014temporal}}}}
    		}child[sibling distance=7mm]{
    			node[termL3]{Sequential: \cnocat{\emph{Rule-based \cite{chen1998multimodal}}}}
    		}
    	}
    };
\end{tikzpicture}}
\centering
\vspace{-0.7cm}
\caption{\scriptsize{Taxonomy for AFER in Computer Vision. \crgb{Red} corresponds to RGB, \cdepth{green} to 3D, and \cthermal{purple} to thermal.}}
\label{fig:taxonomy}
\vspace{-0.5cm}
\end{figure*}

In Figure \ref{fig:taxonomy} we propose a taxonomy for AFER, built along two main components: parametrization and recognition of FEs. These are important components of an automatic FE recognition system, regardless of the data modality.

Parametrization deals with defining coding schemes for describing FEs. Coding schemes may be categorized into two main classes. \emph{Descriptive} coding schemes parametrize FE in terms of surface properties. They focus on what the face can do. \emph{Judgmental} coding schemes describe FEs in terms of the latent emotions or affects that are believed to generate them. Please refer to Section \ref{sec:taxonomy:parameterization} for further details.

An automatic facial analysis system from images or video usually consists of four main parts. First, faces have to be localized in the image (Section \ref{sec:taxonomy:detection:detection}). Second, for many methods a face registration has to be performed. During registration, fiducial points (e.g., the corners of the mouth or the eyes) are detected, allowing for a particularization of the face to different poses and deformations (Section \ref{sec:taxonomy:detection:alignment}). In a third step, features are extracted from the face with techniques dependent on the data modality. A common taxonomy is described for the three considered modalities: RGB, 3D and thermal. The approaches are divided into geometric or appearance based, global or local, and static or dynamic (Section \ref{sec:taxonomy:detection:featureextraction}). Other approaches use a combination of these categories. Finally, machine learning techniques are used to discriminate between FEs. These techniques can predict a categorical expression or represent the expression in a continuous output space, and can model or not temporal information about the dynamics of FEs (Section \ref{sec:taxonomy:detection:learning}).

An additional step, \emph{multimodal fusion} (Section \ref{sec:taxonomy:fusion}), is required when dealing with multiple data modalities, usually coming from other sources of information such as speech and physiological data. This step can be approached in four different ways, depending on the stage at which it is introduced: direct, early, late and sequential fusion.

Modern FE recognition techniques rely on labeled data to learn discriminative patterns for recognition and, in many cases, feature extraction. For this reason we introduce in Section \ref{sec:taxonomy:datasets} the main datasets for all three modalities. These are characterized based on the content of the labeled data, the capture conditions and participants distribution.

\vspace{-0.2cm}

\subsection{Parameterization of FEs}
\label{sec:taxonomy:parameterization}

\begin{figure}[t!]  
    \centering
    \includegraphics[width=\linewidth]{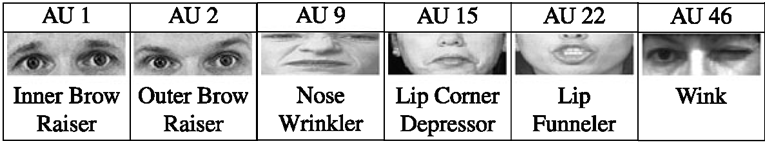}
    \vspace{-0.7cm}\caption{\scriptsize{ \label{fig:facs_aus} Examples of lower and upper face AUs in FACS. Reprinted from \cite{what-when-how}.}}
    \label{fig:aus}
    \vspace{-0.5cm}
\end{figure}

\textbf{Descriptive} coding schemes focus on what the face can do. The most well known examples of such systems are \emph{Facial Action Coding System} (FACS) and \emph{Face Animation Paramters} (FAP). Perhaps the most influential, FACS (1978; 2002) seeks to describe nearly all possible FEs in terms of anatomically-based facial actions \cite{ekman78,ekman02}. The FEs are coded in \emph{Action Units} (AU), which define the contraction of one or more facial muscles (see Figure \ref{fig:facs_aus}). FACS also provides the rules for visual detection of AUs and their temporal segments (onset, apex, offset, ordinal intensity). For relating FEs to emotions, Ekman and Friesen later developed the EMFACS (Emotion FACS), which scores facial actions relevant for particular emotion displays \cite{emfacs}. FAP is now part of the MPEG4 standard and is used for synthesizing FE for animating virtual faces. Is is rarely used to parametrize FEs for recognition purposes \cite{pardas02,aleksic06}. Its coding scheme is based on the position of key feature control points in a mesh model of the face.
\emph{Maximally Discriminative Facial Movement Coding System} (MAX) \cite{izard1983maximally}, another descriptive system, is less granular and less comprehensive. Brow raise in MAX, for instance, corresponds to two separate actions in FACS. It is a truly sign-based approach as it makes no inferences about underlying emotions.

\textbf{Judgmental} coding schemes, on the other hand, describe FEs in terms of the latent emotions or affects that are believed to generate them. Because a single emotion or affect may result in multiple expressions, there is no 1:1 correspondence between what the face does and its emotion label. A hybrid approach is to define emotion labels in terms of specific signs rather than latent emotions or affects. Examples are EMFACS and AFFEX \cite{izard1983system}. In each, expressions related to each emotion are defined descriptively. As an example, enjoyment may be defined by an expression displaying an oblique lip-corner pull co-occurring with cheek raise. Hybrid systems are similar to judgment-based systems in that there is an assumed 1:1 correspondence between emotion labels and signs that describe them. For this reason, we group hybrid approaches with judgment-based systems.
  
\vspace{-0.5cm}

\vspace{-0.2cm} \subsection{Recognition of FEs}
\label{sec:taxonomy:detection}

An AFER system consists of four steps: face detection, face registration, feature extraction and expression recognition.

\vspace{-0.2cm}\subsubsection{Face localization}
\label{sec:taxonomy:detection:detection}

We discuss two main face localization approaches. Detection approaches locate the faces present in the data, obtaining their bounding box or geometry. Segmentation assigns a binary label to each pixel. The reader is referred to \cite{zhang2010survey} for an extensive review on face localization approaches.

For \textbf{RGB images}, Viola\&Jones \cite{viola2001rapid} still is one of the most used algorithms \cite{fea12,salah11,jones2003fast}. It is based on a cascade of weak classifiers, but while fast, it has problems with occlusions and large pose variations \cite{salah11}. Some methods overcome these weaknesses by considering multiple pose-speccific detectors and either a pose router \cite{jones2003fast} or a probabilistic approach \cite{wu2004fast}. Other approaches include \emph{Convolutional Neural Networks} (CNN) \cite{osadchy2007synergistic} and \emph{Support Vector Machines} (SVM) applied over HOG features \cite{dalal2005histograms}. While the later achieves a lower accuracy when compared to Viola\&Jones, the CNN approach in \cite{osadchy2007synergistic} allows for comparable accuracies over wide range of poses.


Regarding face segmentation, early works usually exploit color and texture information along with ellipsoid fitting \cite{sobottka1996segmentation,sirohey1998human,sobottka1998novel}. A posterior step is introduced in \cite{chai1999face} to correct prediction gaps and wrongly labeled background pixels. Some works use segmentation to reduce the search space during face detection \cite{lakshmi2010segmentation}, while others use a \emph{Face Saliency Map} (FSM) \cite{li2008saliency} to fit a geometric model of the face and perform a boundary correction procedure.

For \textbf{3D images} \cite{colombo20063d,nair20093} use curvature features to detect high curvature areas such as the nose tip and eye cavities. Segmentation is also applied to 3D face detection. \cite{pamplona2010automatic} uses k-means to discard the background and locates candidate faces through edge and ellipsoid detection, selecting the highest probability fitting. In \cite{hernandez2012graph}, \emph{Random Forests} are used to assign a body part label to each pixel, including the face. This approach was latter extended in \cite{shotton2013real}, using \emph{Graph Cuts} (GC) to optimize the Random Forest probabilities.

While RGB techniques are applicable to \textbf{thermal images}, segmenting the image according to the \emph{radiant emittance} of each pixel \cite{koda2009facial,trujillo2005automatic} usually is enough. 


\vspace{-0.2cm}\subsubsection{Face registration}
\label{sec:taxonomy:detection:alignment}

\begin{figure}
	\includegraphics[height= 1.8cm, width=\linewidth]{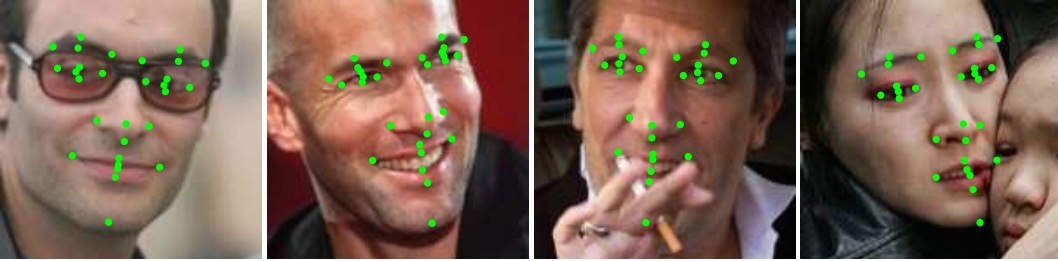}
\vspace{-0.7cm}    \caption{\scriptsize{Sample images from the LFPW dataset aligned with the \emph{Supervised Descent Method} (SDM). Obtained from \cite{xiong2013supervised}.}}
	\label{fig:geom}
\vspace{-0.5cm}\end{figure}

Once the face is detected, fiducial points (\emph{aka.} landmarks) are located (see Figure \ref{fig:geom}). This step is necessary in many AFER approaches in order to rotate or frontalize the face. Equivalently, in the 3D case the face geometry is registered against a 3D geometric model. A thorough review on this subject is out of the scope of this work. The reader is referred to \cite{wang2014facial} and \cite{tam2013registration} for 2D and 3D surveys respectively.

Different approaches are used for grayscale, RGB and near-infrared modalities, and for 3D. In the first case, the objective is to exploit visual information to perform feature detection, a process usually referred to as \emph{landmark localization} or \emph{face alignment}. In the 3D case, the acquired geometry is registered to a shape model through a process known as \emph{face registration}, which minimizes the distance between both. While these processes are distinct, sometimes the same name is used in the literature. To prevent confusion, this work refers to them as 2D and 3D face registration.

\textbf{2D face registration}. \emph{Active Appearance Models} (AAM) \cite{cootes2001active} is one of the most used methods for 2D face registration. It is an extension of \emph{Active Shape Models} (ASM) \cite{cootes1995active} which encodes both geometry and intensity information. 3D versions of AAM have also been proposed \cite{romdhani2003efficient}, but making alignment much slower due to the impossibility of decoupling shape and appearance fitting. This limitation is circumvented in \cite{baker2004lucas}, where a 2D model is fit while a 3D one restricts its shape variations. Another possibility is to generate a 2D model from 3D data through a continuous, uniform sampling of its rotations \cite{igual2014continuous}.

The real-time method of \cite{dantone2012real} uses \emph{Conditional Regression Forests} (CRF) over a dense grid, extracting intensity features and Gabor wavelets at each cell. A more recent set of real-time methods is based on regressing the shape through a cascade of linear regressors. As an example, \emph{Supervised Descent Method} (SDM) \cite{xiong2013supervised} uses simplified SIFT features extracted at each landmark estimate.

\textbf{3D face registration}. In the 3D case, the goal is to find a geometric correspondence between the captured geometry and a model. \emph{Iterative Closest Point} (ICP) \cite{besl1992method} iteratively aligns the closest points of two shapes. In \cite{alyuz2012adaptive}, visible patches of the face are detected and used to discard obstructions before using ICP for alignment. In the case of non-rigid registration, it allows the matched 3D model to deform. In \cite{mao2004constructing}, a correspondence is established manually between landmarks of the model and the captured data, using a \emph{Thin Plate Spline} (TPS) model to deform the shape. \cite{tena2006validated} improves the method by using multi-resolution fitting, an adaptive correspondence search range, and enforcing symmetry constraints. \cite{szeptycki2009coarse} uses a coarse-to-fine approach based on the shape curvature. It initially locates the nose tip and eye cavities, afterwards localizing finer features. Similarly, \cite{alyuz2010regional} first finds the symmetry axis of the face in order to facilitate feature matching. Other techniques include registering a \emph{3D Morphable Model} 3DMM \cite{blanz1999morphable}, 3D-ASM \cite{fanelli2013real} or deformable 2D triangular mesh \cite{savran2008non}, and registering a 3D model through \emph{Simulated Annealing} (SA) \cite{queirolo20103d}.


\vspace{-0.2cm}\subsubsection{Feature extraction}
\label{sec:taxonomy:detection:featureextraction}

Extracted features can be divided into predesigned and learned. Predesigned features are hand-crafted to extract relevant information. Learned features are automatically learned from the training data. This is the case of deep learning approaches, which jointly learn the feature extraction and classficiation/regression weights. These categories are further divided into global and local, where global features extract information from the whole facial region, and local ones from specific regions of interest, usually corresponding to AUs. Features can also be split into static and dynamic, with static features describing a single frame or image and dynamic ones including temporal information.


\textbf{Predesigned features} can also be divided into appearance and geometrical. Appearance features use the intensity information of the image, while geometrical ones measure distances, deformations, curvatures and other geometric properties. This is not the case of learned features, for which the nature of the extracted information is usually unknown. 


Geometric features describe faces through distances and shapes. These cannot be extracted from thermal data, since dull facial features difficult the precise localization of landmarks. Global geometric features, for both RGB and 3D modalities, usually describe the face deformation based on the location of specific fiducial points. For RGB, \cite{pantic06} uses the distance between fiducial points. The deformation parameters of a mesh model are used in \cite{sebe07,kotsia07}. Similarly, for 3D data \cite{tang2008segments} use the distance between pairs of 3D landmarks, while \cite{mpiperis20083d} uses the deformation parameters of an EDM. Manifolds are used in \cite{chang05} to describe the shape deformation of a fitted 3D mesh separately at each frame of a video sequence through Lipschitz embedding.

The use of 3D data allows generating 2D representations of facial geometry such as depth maps \cite{berretti20113d,vretos11}. In \cite{sandbach12lbp} \emph{Local Binary Patterns} (LBP) are computed over different 2D representations, extracting histograms from them. Similarly, \cite{hayat13} uses SVD to extract the 4 principal components from LBP histograms. In \cite{zeng13}, the geometry is described through the \emph{Conformal Factor Image} (CFI) and \emph{Mean Curvature Image} (MCI). \cite{lemaire2013fully} captures the mean curvatures at each location with \emph{Differential Mean Curvature Maps} (DMCM), using HOG histograms to describe the resulting map. 

In the dynamic case the goal is to describe how the face geometry changes over time. For RGB data, facial motions are estimated from color or intensity information, usually through \emph{Optical flow} \cite{wollmer2013lstm}. Other descriptors such as \emph{Motion History Images} (MHI) and \emph{Free-Form Deformations} (FFDs) are also used \cite{koelstra10}. In the 3D case, much denser geometric data facilitates a global description of the facial motions. This is done either through deformation descriptors or motion vectors. \cite{le11} extracts and segments level curvatures, describing the deformation of each segment. FFDs are used in \cite{sandbach11} to register the motion between contiguous frames, extracting features through a quad-tree decomposition. \emph{Flow images} are extracted from contiguous frame pairs in \cite{fang12}, stacking and describing them with LBP-TOP.


In the case of local geometric feature extraction, deformations or motions in localized regions of the face are described. Because these regions are localized, it is difficult to geometrically describe their deformations in the RGB case (being restricted by the precision of the face registration step). As such, most techniques are dynamic for RGB data. In the case of 3D data, where much denser geometric information is available, the opposite happens. 

In the static case, some 3D approaches describe the curvature at specific facial regions, either using primitives \cite{wang20063d} or closed curves \cite{maalej2011shape}. Others describe local deformations through SIFT descriptors \cite{berretti20113d} extracted from the depth map or HOG histograms extracted from DMCM feature maps \cite{lemaire2013fully}. In \cite{gong2009automatic} the \emph{Basic Facial Shape Components} (BFSC) of the neutral face are estimated from the expressive one, subtracting the expressive and neutral face depth maps at rectangular regions around the eyes and mouth.

Most dynamic descriptors in the geometric, local case have been developed for the RGB modality. These are either based on landmark displacements, coded with \emph{Motion Units} \cite{cohen03,cohen03learning}, or the deformation of certain facial components such as the mouth, eyebrows and eyes, coded with FAP \cite{pardas02,aleksic06}. One exception is the work in \cite{yin2006analyzing} over 3D data, where an EDM locates a set of landmarks and a motion vector is extracted from each landmark and pair of frames.


Although geometrical features are effective for describing FEs, they fail to detect subtler characteristics like wrinkles, furrows or skin texture changes. Appearance features are more stable to noise, allowing for the detection of a more complete set of FEs, being particularly important for detecting microexpressions. These feature extraction techniques are applicable to both RGB and thermal modalities, but not to 3D data, which does not convey appearance information.

Global appearance features are based on standard feature descriptors extracted on the whole facial region. For RGB data, usually these descriptors are applied either over the whole facial patch or at each cell of a grid. Some examples include \emph{Gabor filters} \cite{littlewort04,littlewort2011computer}, LBP \cite{shan09,savran2014temporal}, \emph{Pyramids of Histograms of Gradients} (PHOG) \cite{sun2014combining,dhall2011emotion}, \emph{Multi-Scale Dense SIFT} (MSDF) \cite{sun2014combining} and \emph{Local Phase Quantization} (LPQ) \cite{dhall2011emotion}. In \cite{lyons99} a grid is deformed to match the face geometry, afterwards applying \emph{Gabor filters} at each vertex. In \cite{gu2012facial} the facial region is divided by a grid, applying a bank of \emph{Gabor filters} at each cell and radially encoding the mean intensity of each feature map. An approach called \emph{Graph-Preserving Sparse Non-negative Matrix Factorization} (GSNMF) \cite{zhi2011graph} finds the closest match to a set of base images and assigns its associated primary emotion. This approach is improved in \cite{zafeiriou2009nonlinear}, where \emph{Projected Gradient Kernel Non-negative Matrix Factorization} (PGKNMF) is proposed.

In the case of thermal images, the dullness of the image makes it difficult to exploit the facial geometry. This means that, in the global case, the whole facial patch is used. The descriptors exploit the difference of temperature between regions. One of the first works \cite{yoshitomi1997facial} generated a series of \emph{Binary Differential Images} (BDI), extracting the ratio of positive area divided by the mean ratio over the training samples. \emph{2D Discrete Cosine Transform} (2D-DCT) is used in \cite{koda2009facial,yoshitomi2010facial} to decompose the frontalized face into cosine waves, from which an heuristic approach extracts features.

Dynamic global appearance descriptors are extensions to 3 dimensions of the already explained static global descriptors. For instance, \emph{Local Binary Pattern histograms from Three Orthogonal Planes} (LBP-TOP) are used for RGB data \cite{zhao2007dynamic}. LBP-TOP is an extension of LBP computed over three orthogonal planes at each bin of a 3D volume formed by stacking the frames. \cite{sun2014combining} uses a combination of LBP-TOP and \emph{Local Phase Quantization from Three Orthogonal Planes} (LPQ-TOP), a descriptor similar to LBP-TOP but more robust to blur. LPQ-TOP is also used in \cite{he2015multimodal}, along with \emph{Local Gabor Binary Patterns from Three Orthogonal Planes} (LGBP-TOP). In \cite{liu2014combining}, a combination of HOG, SIFT and CNN are extracted at each frame. The first two are extracted from an overlapping grid, while the CNN extracts features from the whole facial patch. These are evaluated independently over time and embedded into Riemannian manifolds. For thermal images, \cite{liu2011emotion} uses a combination of \emph{Temperature Difference Histogram Features} (TDHFs) and \emph{Thermal Statistic features} (StaFs). TDHFs consist of histograms extracted over the difference of thermal images. StaFs are a series of 5 basic statistical measures extracted from the same difference images.


Local appearance features are not used as frequently as global ones, since it requires previous knowledge to determine the regions of interest. In spite of that, some works use them for both RGB and thermal modalities. In the case of static features, \cite{geetha2009facial} describes the appearance of grayscale frames by spreading an array of cells across the mouth and extracting the \emph{mean intensity} from each. For thermal images, \cite{trujillo2005automatic} generates eigenimages from each region of interest and uses the principal component values as features. In \cite{hernandez2007visual} \emph{Gray Level Co-occurrence Maxrices} (GLCMs) are extracted from the interest regions and second-order statistics computed on them. GLCM encode texture information by representing the occurrence frequencies of pairs of pixel intensities at a given distance. As such, these are also applicable to the RGB case. In \cite{wang2014emotion} a combination of StaFs, 2D-DCT and GLCM features is used, extracting both local and global information.

Few works consider dynamic local appearance features. The only one to our knowledge \cite{liu2015spontaneous} describes thermal sequences by processing them with \emph{SIFT flow} and chunking them into clips. Contiguous clip frames are wrapped and subtracted, spatially dividing the clip with a grid. The resulting cuboids with higher inter-frame variability for either radiance or flow are selected, extracting a \emph{Bag of Words histogram} (BoW Hist.) from each.


Based on the observation that some AU are better detected using geometrical features and others using appearance ones, it was suggested that a combination of both might increase recognition performance \cite{tian01,pantic07,koelstra10}. Feature extraction methods combining geometry and appearance are more common for RGB, but it is also possible to combine RGB and 3D. Because 3D data is highly discriminative and robust to problems such as shadows and illumination changes, the benefits of combining it with RGB data are small. Nevertheless, some works have done so \cite{ramanathan2006human,zhao2010automatic,zhao2013unified}. It should also be possible to extract features combining 3D and thermal information, but to the best of our knowledge it has not been attempted.

In the static case, \cite{tian01} uses a combination of Multi-state models and edge detection to detect 18 different AUs on the upper and lower parts of the face in grayscale images. \cite{dapogny2015dynamic} uses both global geometry features and local appearance features, combining landmark distances and angles with HOG histograms centered at the barycenter of triangles specified by three landmarks. Other approaches use deformable models such as 3DMM \cite{ramanathan2006human} to combine 3D and intensity information. In \cite{zhao2010automatic,zhao2013unified} SFAM describes the deformation of a set of distance-based, patch-based and grayscale appearance features encoded using LBP.

When analysing dynamic information, \cite{dapogny2015dynamic} uses RGB data to combine the landmark displacements between two frames with the change in intensity of pixels located at the barycenter defined by three landmarks.


\textbf{Learned} features are usually trained through a joint feature learning and classification pipeline. As such, these methods are explained in Section \ref{sec:taxonomy:detection:learning} along with learning. The resulting features usually cannot be classified as local or global. For instance, in the case of CNNs, multiple convolution and pooling layers may lead to higher-level features comprising the whole face, or to a pool of local features. This may happen implicitly, due to the complexity of the problem, or by design, due to the topology of the network. In other cases, this locality may be hand-crafted by restricting the input data. For instance, the method in \cite{liu2014facial}, selects interest regions and describes each one with a \emph{Deep Belief Network} (DBN). Each DBN is jointly trained with a weak classifier in a boosted approach.

\subsubsection {FE classification and regression}
\label{sec:taxonomy:detection:learning}

\begin{figure*}
    \centering
    \includegraphics[width=15cm]{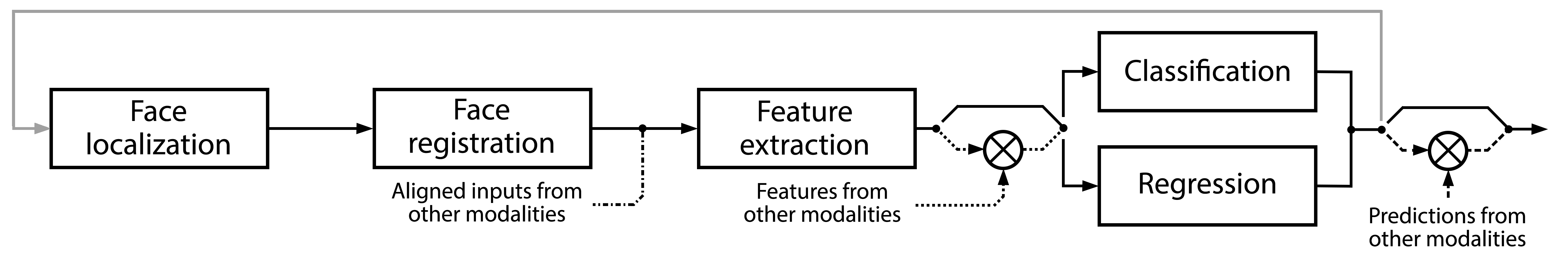}
    \vspace{-0.3cm}\caption{General execution pipeline for the different modality fusion approaches. The tensor product symbols represent the modality fusion strategy. Approach-specific components of the pipeline are represented with different line types: dotted corresponds to early fusion, dashed to late fusion, dashed-dotted to direct data fusion and gray to sequential fusion.}
    \label{fig:fusion}
\vspace{-0.5cm}\end{figure*}

FE recognition techniques are grouped into categorical and continuous depending on the target expressions \cite{martinez12}. In the categorical case there is a predefined set of expressions. Commonly for each one a classifier is trained, although other ensemble strategies could be applied. Some works detect the six primary expressions \cite{littlewort04,kotsia07,sebe07}, while others detect expressions of pain, drowsiness and emotional attachment \cite{ashraf09,littlewort09,lucey10}, or indices of psychiatric disorder \cite{cohn09,kohler08}.

In the continuous case, FEs are represented as points in a continuous multidimensional space \cite{zeng09}. The advantages of this second approach are the ability to represent subtly different expressions, mixtures of primary expressions, and the ability to unsupervisedly define the expressions through clustering. Many continuous models are based on the activation-evaluation space. In \cite{caridakis2006modeling}, a \emph{Recurrent Neural Network} (RNN) is trained to predict the real-valued position of an expression inside that space. In \cite{nicolle2012robust} the feature space is scaled according to the correlation between features and target dimensions, clustering the data and performing \emph{Kernel regression}. In other cases like \cite{fragopanagos2005emotion}, which uses a RNN for classification, each quadrant is considered as a class, along with a fifth neutral target.

Expression recognition methods can also be grouped into static and dynamic. Static models evaluate each frame independently, using classification techniques such as \emph{Bayesian Network Classifiers} (BNC) \cite{cohen03,cohen03learning,sebe07}, \emph{Neural Networks} (NN) \cite{yoshitomi1997facial,tian01}, \emph{Support Vector Machines} (SVM) \cite{littlewort04,kotsia07,berretti20113d,lemaire2013fully,trujillo2005automatic}, SVM committees \cite{hernandez2007visual} and \emph{Random Forests} (RF) \cite{dapogny2015dynamic}. In \cite{gu2012facial} \emph{k-Nearest Neighbors} (kNN) is used to separately classify local patches, performing a dimensionality reduction of the outputs through PCA and LDA and classifying the resulting feature vector.

More recently, deep learning architectures have been used to jointly perform feature extraction and recognition. These approaches often use pre-training \cite{hinton06}, an unsupervised layer-wise training step that allows for much larger, unlabeled datasets to be used. CNNs are used in \cite{ranzato11,rifai12,liu2014learning,kahou2013combining,song2014deep}. \cite{liu2013aware} proposes \emph{AU-aware Deep Networks} (AUDN), where a common convolutional plus pooling step extracts an over-complete representation of expression features, from which receptive fields map the relevant features for each expression. Each receptive field is fed to a DBN to obtain a non-linear feature representation, using an SVM to detect each expression independently. In \cite{liu2014facial} a two-step iterative process is used to train \emph{Boosted Deep Belief Networks} (BDBN) where eacn DBN learns a non-linear feature from a face patch, jointly performing feature learning, selection and classifier training. \cite{he2013facial} uses a \emph{Deep Boltzmann Machine} (DBM) to detect FEs from thermal images. Regarding 3D data, \cite{ijjina2014facial} transforms the facial depth map into a gradient orientation map and performs classification using a CNN.

Dynamic models take into account features extracted independently from each frame to model the evolution of the expression over time. Dynamic Bayesian Networks such as \emph{Hidden Markov Models} (HMM) \cite{cohen03,aleksic06,koelstra10,le11,sandbach11,wu2015multi,pardas02} and \emph{Variable-State Latent Conditional Random Fields} (VSL-CRF) \cite{walecki15} are used. Other techniques use RNN architectures such as \emph{Long Short Term Memory} (LSTM) networks \cite{wollmer2013lstm}. In other cases \cite{tsalakanidou2010real,tsalakanidou2009robust}, hand-crafted rules are used to evaluate the current frame expression against a reference frame. In \cite{dapogny2015dynamic} the transition probabilities between FEs given two frames are first evaluated with RF. The average of the transition probabilities from previous frames to the current one, and the probability for each expression given the individual frame are averaged to predict the final expression. Other approaches classify each frame independently (\emph{eg.} with SVM classifiers \cite{geetha2009facial}), using the prediction averages to determine the final FE.

In \cite{sebe07,fang12} an intermediate approach is proposed where motion features between contiguous frames are extracted from interest regions, afterwards using static classification techniques. \cite{liu2014combining} encodes statistical information of frame-level features into Riemannian manifolds, and evaluates three approaches to classify the FEs: SVM, \emph{Logistic regression} (LR) and \emph{Partial Least Squares} (PLS).

More redently, dynamic, continuous models have also been considered. \emph{Deep Bidirectional Long Short-Term Memory Recurrent Neural Networks} (DBLSTM-RNN) are used in \cite{he2015multimodal}. While \cite{savran2012combining} uses static methods to make the initial affect pedictions at each time step, it uses particle filters to make the final prediction. This both reduces noise and performs modality fusion.

\subsubsection {Multimodal fusion techniques}
\label{sec:taxonomy:fusion}

\begin{figure*}[t]
	\centering
	\begin{tabular}{ccc}
      \subfloat[CK\&CK+]{\includegraphics[width = 5.7cm,height=2cm]{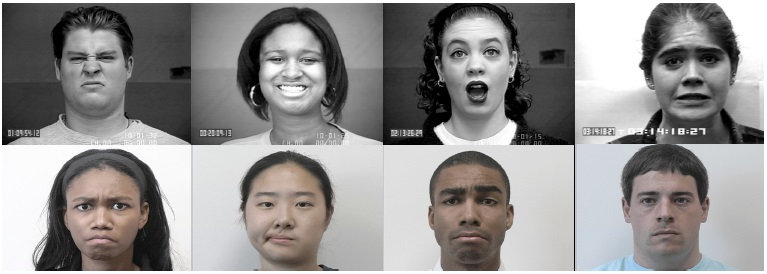}} &
      \subfloat[MMI]{\includegraphics[width = 5.7cm,height=2cm]{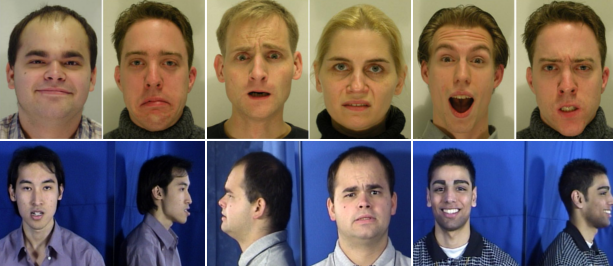}} &
      \subfloat[Multi-PIE]{\includegraphics[width = 5.7cm,height=2cm]{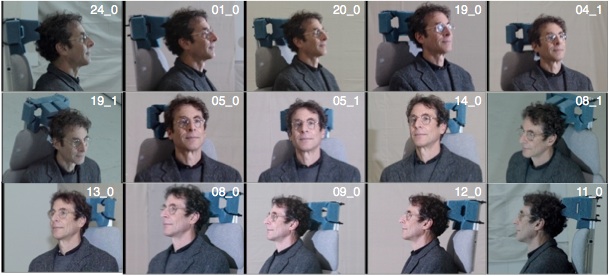}} \\
      \subfloat[SFEW]{\includegraphics[width = 5.7cm,height=2.1cm]{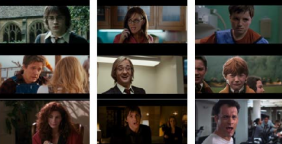}} &
      \subfloat[Bosphorus]{\includegraphics[width = 5.7cm,height=2.1cm]{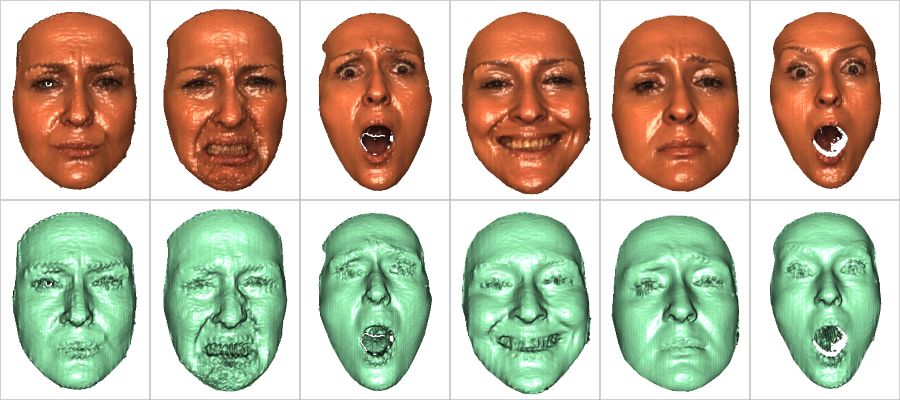}} &
      \subfloat[KTFE]{\includegraphics[width = 5.7cm,height=2.1cm]{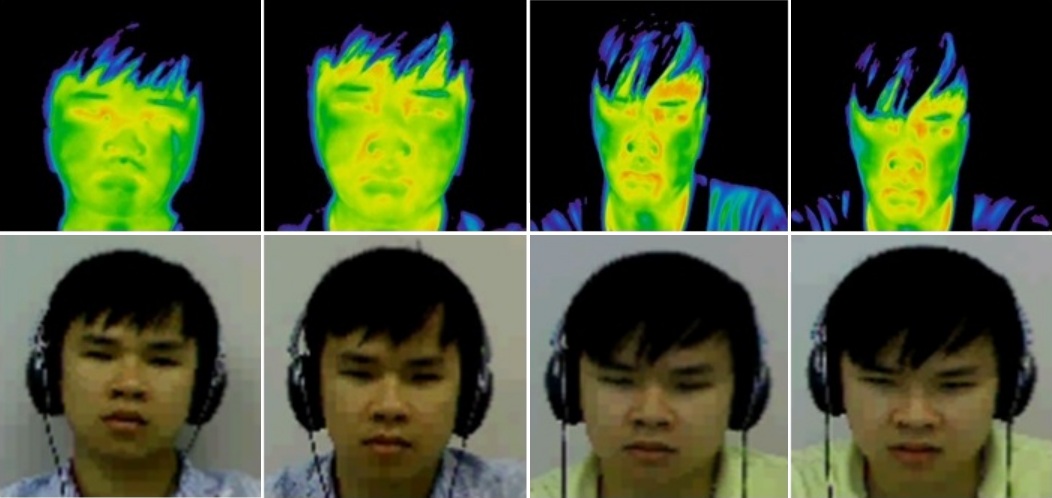}}
	\end{tabular}

\vspace{-0.3cm}	\caption{\scriptsize{FE datasets. (a) The CK \cite{kanade00} dataset (top) contains posed exaggerated expressions. The CK+ \cite{ck+} (bottom) extends CK by introducing spontaneous expressions. (b) MMI \cite{mmi}, the first dataset to contain profile views. (c) MultiPIE \cite{multi-pie} has multiview samples under varying illumination conditions. (d) SFEW \cite{dhall2011static}, an in the wild dataset. (e) Primary FEs in Bosphorus \cite{bosphorus}, a 3D dataset. (f) KTFE \cite{nguyen2014thermal} dataset, thermal images of primary spontaneous FEs.}}
	\label{fig:datasets}
\vspace{-0.5cm}\end{figure*}

Many works have considered multimodality for recognizing emotions, either by considering different visual modalities describing the face or, more commonly, by using other sources of information (e.g. audio or physiological data). Fusing multiple modalities has the advantage of increased robustness and conveying complementary information. Depth information is robust to changes in illumination, while thermal images convey information related to changes in the blood flow produced by emotions. It has been found that momentary stress increases the periorbital blood flow, while if sustained the blood flow to the forehead increases \cite{pavlidis2007interacting}. Joy decreases the blood flow to the nose, while arousal increases it to the nose, periorbital, lips and forehead \cite{ioannou2014thermal}.

The fusion approaches followed by these works can be grouped into three main categories: \emph{early}, \emph{late} and \emph{sequential fusion} (see Figure \ref{fig:fusion}). Early fusion merges the modalities at the feature level, while late fusion does so after applying expression recognition, at the decision level \cite{wu1999multimodal}. Early fusion directly exploits correlations between features from different modalities, and is specially useful when sources are synchronous in time. However, it forces the classifier/regressor to work with a higher-dimensional feature space, increasing the likelihood of over-fitting. On the other hand, late fusion is usually considered for asynchronous data sources, and can be trained on modality-specific datasets, increasing the amount of available data. A sequential use of modalities is also considered by some multimodal approaches \cite{chen1998multimodal}.

It is also possible to directly merge the input data from different modalities, an approach referred in this document as \emph{direct data fusion}. This approach has the advantage of allowing the extraction of features from a richer data source, but is limited to input data correlated for both spatial and, if considered, temporal domains.


Regarding \textbf{early fusion}, the simplest approach is \emph{plain early fusion}, which consists on concatenating the feature vectors from both modalities. This is done in \cite{huang1998bimodal,wollmer2013lstm} to fuse RGB video and speech. Usually, a feature selection approach is applied. One such technique is \emph{Sequential Backward Selection} (SBS), where the least significant feature is iteratively removed until some criterion is met. In \cite{busso2004analysis} SBS is used to merge RGB video and speech. A more complex approach is to use the \emph{best-first search} algorithm, as done in \cite{gunes2005affect} to fuse RGB facial and body gesture information. Other approaches include using \emph{10-fold cross-validation} to evaluate different subsets of features \cite{kessous2010multimodal} and an \emph{Analysis of Variance} (ANOVA) \cite{d2010multimodal} to independently evaluate the discriminative power of each feature. These two works both fuse RGB video, gesture and speech information.

An alternative to feature selection is to encode the dependencies between features. This can be done by using probabilistic inference models for recognition. A \emph{Bayesian Network} is used in \cite{sebe2006emotion} to infer the emotional state from both RGB video and speech. In \cite{zeng2008audio} a \emph{Multi-stream fused HMM} (MFHMM) models synchronous information on both modalities, taking into account the temporal component. The advantage of probabilistic inference models is that the relations between features are restricted, reducing the degrees of freedom of the model. On the other hand, it also means that it s necessary to manually design these relations. Other inference techniques are also used, such as \emph{Fuzzy Inference Systems} (FIS), to represent emotions in a continuous 4-dimensional output space based on grayscale video, audio and contextual information \cite{soladie2012multimodal}.


\textbf{Late fusion} merges the results of multiple classifiers/regressors into a final prediction. The goal is either to obtain a final class prediction, a continuous output specifying the intensity/confidence for each expression or a continuous value for each dimension in the case of continuous representations. Here the most common late fusion strategies used for emotion recognition are discussed, but since it can be seen as an ensemble learning approach, many other machine learning techniques could be used. The simplest approach is the \emph{Maximum rule} \footnote{Also known as the \emph{ winner takes it all} rule}, which selects the maximum of all posterior probabilities. This is done in \cite{busso2004analysis} to fuse RGB video and speech. This technique is sensible to high-confidence errors. A classifier incorrectly predicting a class with high confidence would be frequently selected as winner even if all other classifiers disagree. This can be partially offset by using a combination of responses, as is the case of the \emph{Sum rule} and \emph{Product rule}. The \emph{Sum rule} sums the confidences for a given class from each classifier, giving the class with the highest confidence as result \cite{gunes2005affect, liu2014combining, busso2004analysis}. The \emph{Product rule} works similarly, but multiplying the confidences \cite{gunes2005affect, busso2004analysis}. While these approaches partially offset the single-classifier weakness problem, the strengths of each individual modality are not considered. The \emph{Weight criterion} solves this by assigning a confidence to each classifier output, otputting a weighted linear combination of the predictions \cite{de1997facial, gunes2005affect, busso2004analysis, yoshitomi2000effect}. A \emph{rule-based} approach is also possible, where a dominant modality is selected for each target class \cite{de2000bimodal}.

\emph{Bayesian Inference} is used to fuse predictions of RGB, speech and lexical classifiers, simultaneously modeling time \cite{savran2014temporal}. The bayesian framework uses information from previous frames along with the predictions from each modality to estimate the emotion displayed at the current frame.


\textbf{Sequential} fusion is a technique that applies the different modality predictions in sequential order. It uses the results of one modality to disambiguate those of another when needed. Few works use this technique, being an example \cite{chen1998multimodal}, a rule-based approach that combines grayscale facial and speech information. The method uses acoustic data to distinguish candidate emotions, disambiguating the results with grayscale information.

\vspace{-0.2cm}\subsection{FE datasets}
\label{sec:taxonomy:datasets}
\begin{figure*}\center	\includegraphics[width=16cm,height=3.5cm]{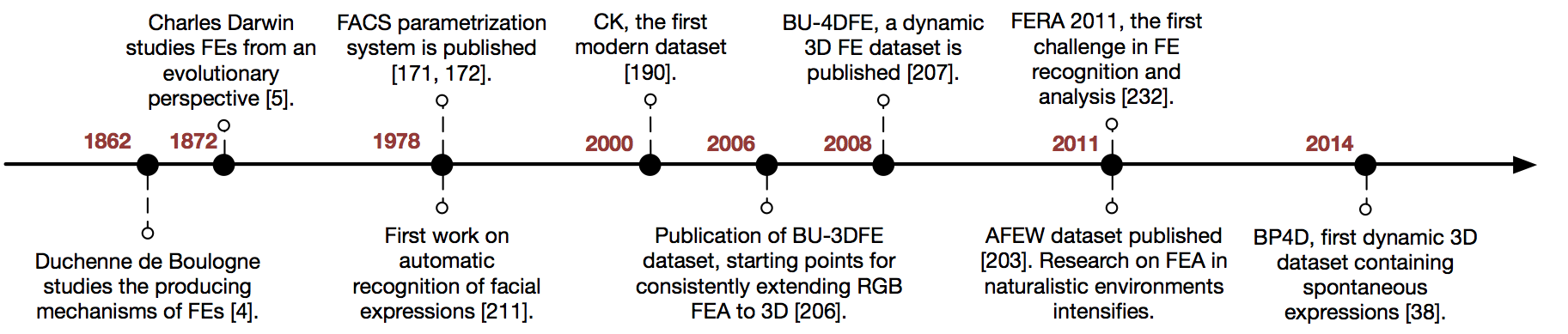}
   \caption{ \scriptsize{ Historical evolution of AFER.}} 
	\label{fig:history}
\vspace{-0.1cm}\end{figure*}

We group datasets' properties in three main categories, focusing on content, capture modality and participants. In the content category we refer to the type of content and labels the datasets provide. We signal intentionality of the FEs (posed or spontaneous), the labels (primary expressions, AUs or others where is the case) and if datasets contain still images or video sequences (static/dynamic). In the capture category we group datasets by the context in which data was captured (lab or non-lab) and diversity in perspective, illumination and occlusions. The last section compiles statistical data about participants, including age, gender and ethnic diversity. In Figure \ref{fig:datasets} we show samples from some of the most well-known datasets. In Tables \ref{tab:db_rgb} and \ref{tab:db_3d_t} the reader can refer to a complete list of RGB, 3D and Thermal datasets and their characteristics.   

\textbf{RGB.} One of the first important datasets made public was the Cohn-Kanade (CK) \cite{kanade00}, later extended into what was called the CK+ \cite{ck+}. The first version is relatively small, consisting of posed primary FEs. It has limited gender, age and ethnic diversity and contains only frontal views with homogeneous illumination. In CK+, the number of posed samples was increased by 22\% and spontaneous expressions were added. The MMI dataset was a major improvement \cite{pantic06}. It adds profile views of not only the primary expressions but most of the AU of the FACS system. It also introduced temporal labeling of onset, apex and offset. Multi-PIE \cite{multi-pie} increases the variability by including a very large number of views at different angles and diverse illumination conditions. GEMEP-FERA is a subset of the emotion portrayal dataset GEMEP, specially annotated using FACS. CASME \cite{casme} is an example of a dataset containing microexpressions. A limitation of most RGB datasets is the lack of intensity labels. It is not the case of the DISFA dataset \cite{disfa}. Participants were recorded while watching a video specially chosen for inducing emotional states and 12 AUs were coded for each video frame on a 0 (not present) to 5 (maximum intensity) scale\cite{disfa}.

While previous RGB datasets record FEs in controlled lab environments, \emph{Acted Facial Expressions In The Wild Database} (AFEW) \cite{afew}, \emph{Affectiva-MIT Facial Expression Dataset} (AMFED) \cite{mcduff2013} and SEMAINE \cite{mckeown2012semaine} contain faces in naturalistic environments. AFEW has 957 videos extracted from movies, labeled with six primary expressions and additional information about pose, age, and gender of multiple persons in a frame.  AMFED contains spontaneous FEs recorded in natural settings over the Internet. Metadata consists of frame by frame AU labelling and self reporting of affective states. SEMAINE contains primitive FEs, FACS annotations, labels of cognitive states, laughs, nods and shakes during interactions with artificial agents.

\begin{table*}
\centering
\caption{\label{tab:db_rgb} A non-comprehensive list of RGB FE datasets.}\vspace{-0.3cm}
\resizebox{\textwidth}{!}{%
\begin{tabular}{| c | c  c || c | c | c | c | c | c | c | c | c | c | c |}             
\hline
  &  &  & \multicolumn{11}{c|}{\textbf{RGB}}  \\ \hline 
  &  &  & \textbf{CK+} & \textbf{MPIE} & \textbf{JAFFE} & \textbf{MMI} & \textbf{RU\_FACS} & \textbf{SEMAINE} & \textbf{CASME} & \textbf{DISFA} & \textbf{AFEW} & \textbf{SFEW} & \textbf{AMFED} \\ \hline \hline

  \parbox[t]{2mm}{\multirow{3}{*}{\rotatebox[origin=c]{90}{\textbf{\scriptsize{Content}}}}} & \multicolumn{2}{c||}{Intention(Posed/Spontaneous)} & $P$ & $P$ & $P$ & $P$ & $S$ & $S$ & $S$ & $S$ & $S$ & $S$ & $S$ \\ \cline{2-14}
  & \multicolumn{2}{c||}{Label(Primary/AU/DA)} & $P/AU$ & $P$ & $P$ & $AU+T$ & $P/AU$ & $P/AU/DA^1$ & $P/AU$ & $AU+I$ &$P/^2$ & $P$ & $P/AU/Smile$ \\ \cline{2-14}
  & \multicolumn{2}{c||}{Temporality(Static/Dynamic)} & $D$ & $S$ & $S$ & $D$ & $D$ & $D$ & $D$ & $D$ & $D$ & $S$ & $D$ \\ \hline \hline
  
  \parbox[t]{2mm}{\multirow{4}{*}{\rotatebox[origin=c]{90}{\textbf{\scriptsize{Capture}}}}} & \multicolumn{2}{c||}{Environment(Lab/Non-lab)} & $L$ & $L$ & $L$ & $L$ & $L$ & $L$ & $L$ & $L$ & $N$ & $N$ & $N$ \\ \cline{2-14}
  & \multicolumn{2}{c||}{Multiple Perspective} & $\circ$ & \textbullet & $\circ$ &  \textbullet & \textbullet & $\circ$ & $\circ$ & $\circ$ & \textbullet & \textbullet & \textbullet \\ \cline{2-14} 
  & \multicolumn{2}{c||}{Multiple Illumination}  & $\circ$ & \textbullet & $\circ$ & \textbullet & $\circ$ & $\circ$ & $\circ$ & $\circ$ & \textbullet & \textbullet & \textbullet \\ \cline{2-14} 
  & \multicolumn{2}{c ||}{Occlusions}  & $\circ$ & \textbullet & $\circ$ &  $\circ$ & \textbullet & $\circ$ & $\circ$ & $\circ$ & \textbullet & \textbullet & $\circ$ \\ \hline \hline 
  
  \parbox[t]{2mm}{\multirow{4}{*}{\rotatebox[origin=c]{90}{\textbf{\scriptsize{Subjects}}}}} & \multicolumn{2}{c||}{\# of subjects} & 201 & 337 & 10 & 75 & 100 & 150 & 35 & 27 & 220 & 68 & 5268 \\ \cline{2-14}
  & \multicolumn{2}{c||}{Ethnic Diverse} & \textbullet & \textbullet & $\circ$ & \textbullet & $\circ$ & $\circ$ & $\circ$ & \textbullet & \textbullet & \textbullet & \textbullet \\ \cline{2-14}
  & \multicolumn{2}{c||}{Gender(Male/Female(\%))} & 31/69 & 70/30 & 100/0 & 50/50 & - & 62/38 & 37/63 & 44/56 & - & - & 58/42 \\ \cline{2-14} 
  & \multicolumn{2}{  c ||}{Age} & 18-50 & $\mu=27.9$ & - & 19-62 & 18-30 & 22-60 & $\mu=22$ & 18-50 & 1-70 & - & - \\ \hline 
  
  \hline    
  \end{tabular}}
  
  \scriptsize{\textbullet = Yes, $\circ$ = No, - = Not enough information. DA: Dimensional Affect, I = Intensity labelling, T = Temporal segments. $^1$ Other labels include Laughs, Nods, Epistemic states(e.g. Certain, Agreeing, Interested etc.) etc. Refer to original paper for details \cite{mckeown2012semaine}. $^2$ Pose, Age, Gender}. Refer to original paper for details \cite{afew}.
\vspace{-0.1cm}\end{table*}

\begin{table*}
\centering
\caption{\label{tab:db_3d_t} A non-comprehensive list of 3D and Thermal FE datasets.}\vspace{-0.3cm}
\resizebox{0.7\textwidth}{!}{%
\begin{tabular}{| c | c  c || c | c | c | c | c | c | c | c |}             
\hline

  &  &  & \multicolumn{4}{c|}{\textbf{3D}} & \multicolumn{4}{c|}{\textbf{RGB+Thermal}}  \\ \hline 
  &  &  & \textbf{BU-3DFE} & \textbf{BU-4DFE} & \textbf{Bosphorus} & \textbf{BP4D} & \textbf{IRIS} & \textbf{NIST} & \textbf{NVIE} & \textbf{KTFE} \\ \hline \hline

  \parbox[t]{2mm}{\multirow{3}{*}{\rotatebox[origin=c]{90}{\textbf{\scriptsize{Content}}}}} & \multicolumn{2}{c||}{Intention(Posed/Spontaneous)} & $P$ & $P$ & $P$ & $S$ & $P$ & $P$ & $S/P$ & $S/P$ \\ \cline{2-11}
  & \multicolumn{2}{c||}{Label(Primary/AU)} & $P+I$ & $P$ & $P/AU$ & $AU$ & $P$ & $P$ & $P$ & $P$\\ \cline{2-11}
  & \multicolumn{2}{c||}{Temporality(Static/Dynamic)} & $S$ & $D$ & $S$ & $D$ & $S$ & $S$ & $D$ & $D$ \\ \hline \hline
  
  \parbox[t]{2mm}{\multirow{4}{*}{\rotatebox[origin=c]{90}{\textbf{\scriptsize{Capture}}}}}& \multicolumn{2}{c||}{Environment(Lab/Non-lab)} & $L$ & $L$ & $L$ & $L$ & $L$ & $L$ & $L$ & $L$ \\ \cline{2-11} 
  & \multicolumn{2}{c||}{Multiple Perspective} & \textbullet & \textbullet & - &  \textbullet & \textbullet & \textbullet & \textbullet & \textbullet \\ \cline{2-11} 
  & \multicolumn{2}{c||}{Multiple Illumination}  & $\circ$ & $\circ$ & $\circ$ & $\circ$ & \textbullet & \textbullet & \textbullet & \textbullet \\ \cline{2-11} 
  & \multicolumn{2}{c ||}{Occlusions}  & \textbullet & $\circ$ & \textbullet &  $\circ$ & \textbullet & \textbullet & \textbullet & \textbullet \\ \hline \hline 
  
  \parbox[t]{2mm}{\multirow{4}{*}{\rotatebox[origin=c]{90}{\textbf{\scriptsize{Subjects}}}}} & \multicolumn{2}{c||}{\# of subjects} & 100 & 101 & 105 & 41 & 30 & 90 & 215 & 26 \\ \cline{2-11}
  & \multicolumn{2}{c||}{Ethnic Diverse} & \textbullet & \textbullet & $\circ$ & \textbullet & \textbullet & - & $\circ$ & $\circ$ \\ \cline{2-11}
  & \multicolumn{2}{c||}{Gender(Male/Female(\%))} & 56/44 & 57/43 & 43/57 & 56/44 & - & - & 27/73 & 38/62 \\ \cline{2-11} 
  & \multicolumn{2}{c||}{Age} & 18-70 & 18-45 & 25-35 & 18-29 & - & - & 17-31 & 12-32 \\ \hline 
  
  \hline    
  \end{tabular}}  
  
  \scriptsize{\textbullet = Yes, $\circ$ = No, - = Not enough information, I = Intensity labelling.}
\vspace{-0.5cm}\end{table*}

\textbf{3D.} The most well known 3D datasets are BU-3DFE \cite{bu-3dfe}, Bosphorus \cite{bosphorus} (still images), BU-4DFE \cite{bu-4dfe} (video) and BP4D \cite{bp4d} (video). In BU-3DFE, 6 expression from 100 different subjects are captured on four different intensity levels. Bosphorus has low ethnic diversity but it contains a much larger number of expressions, different head poses and deliberate occlusions. BU-4DFE is a high-resolution 3D dynamic FE dataset \cite{bu-4dfe}. Video sequences, having 100 frames each, are captured from 101 subjects. It only contains primary expressions. BU-3DFE, BU-4DFE and Bosphorus all contain posed expressions. BP4D tries to address this issue with authentic emotion induction tasks \cite{bp4d}. Games, film clips and a cold pressor test for pain elicitation were used to obtain spontaneous FEs. Experienced FACS coders annotated the videos, which were double-checked by the subject's self-report, FACS analysis and human observer ratings \cite{bp4d}.

\textbf{Thermal.} There are few thermal FE datasets, and all of them also include RGB data. The first ones, IRIS \cite{irisdb} and NIST/Equinox \cite{nistdb}, consist of image pairs labeled with three posed primary emotions under various illuminations and head poses. Recently the number of labeled FEs has increased, also including image sequences. The \emph{Natural Visible and Infrared facial Expression database} (NVIE) contains 215 subjects, each displaying six expressions, both spontaneous and posed \cite{wang2010natural}. The spontaneous expressions are triggered through audiovisual media, but not all of them are present for each subject. In the \emph{Kotani Thermal Facial Emotion} (KTFE) dataset subjects display posed and spontaneous motions, also triggered through audiovisual media \cite{nguyen2014thermal}.

\vspace{-0.3cm}\section{Historical evolution and current trends}
\label{sec:methods}

\subsection{Historical evolution}

The first work on AFER was published in 1978 \cite{suwa78}. It was tracking the motion of landmarks in an image sequence. Mostly because of poor face detection and face registration algorithms and limited computational power, the subject received little attention throughout the next decade. The work of Mase and Pentland and Paul Ekman marked a revival of this research topic at the beginning of the nineties \cite{mase91, ekman1993final}. The interested reader can refer to some influential surveys of these early works \cite{samal92, pantic00, fasel03}.

In 2000, the CK dataset was published marking the beginning of modern AFER \cite{tian01}. While a large number of approaches aimed at detecting primary FEs or a limited set of FACS AUs \cite{cohen03, aleksic06, kotsia07, littlewort04}, others focused on a larger set of AUs \cite{tian01, pantic06, koelstra10}. Most of these early works used geometric representations, like vectors for describing the motion of the face \cite{cohen03}, active contours for describing the shape of the mouth and eyebrows \cite{aleksic06}, or deformable 2D mesh models \cite{kotsia07}. Others focused on appearance representations like Gabor filters \cite{littlewort04}, optical flow and LBPs \cite{shan09} or combinations between the two \cite{tian01}. The publication of the BU-3DFE dataset \cite{bu-3dfe} was a starting point for consistently extending RGB FE recognition to 3D. While some of the methods require manual labelling of fiducial vertices during training and testing \cite{wang20063d, soyel2007facial,tang2008automatic}, others are fully automatic \cite{lemaire2013fully, gong2009automatic, vretos11, zeng13}. Most use geometric representations of the 3D faces, like principal directions of surface curvatures to obtain robustness to head rotations \cite{wang20063d}, and normalized Euclidean distances between fiducial points in the 3D space \cite{tang2008automatic}. Some encode global deformations of facial surface (depth differences between a basic facial shape component and an expressional shape component) \cite{gong2009automatic} or local shape representations \cite{sandbach12lbp}. Most of them target primary expressions \cite{wang20063d} but studies about AUs were published as well \cite{savran2012comparative, sandbach12lbp}.  

In the first part of the decade static representations were the primary choice in both RGB \cite{tian01, littlewort04}, 3D \cite{wang20063d, wang05, tang2008automatic, gong2009automatic, berretti20113d, lemaire2013fully} and thermal \cite{hernandez2007visual}. In later years various ways of dynamic representation were also explored like tracking geometrical deformations across frames in RGB \cite{kotsia07, pantic06} and 3D \cite{chang05, le11} or directly extracting features from RGB \cite{koelstra10} and thermal frame sequences \cite{nguyen2014thermal,wang2010natural}.

Besides extended work on improving recognition of posed FEs and AUs, studies on expressions in ever more complex contexts were published. Works on spontaneous facial expression detection \cite{lucey2007investigating, sebe07, valstar2006spontaneous, zeng2006spontaneous}, analysis of complex mental states \cite{el2005real}, detection of fatigue \cite{ji2006aprobabilistic}, frustration \cite{kapoor2007automatic}, pain \cite{littlewort2007faces, ashraf09, littlewort09}, severity of depression \cite{girard2014nonverbal} and psychological distress \cite{lucas2015towards}, and including AFER capabilities in intelligent virtual agents \cite{devault2014simsensei} opened new territory in AFER research.

In summary, research in automatic AFER started at the end of the 1970's, but for more than a decade progress was slow mainly because of limitations of face detection and face registration algorithms and lack of sufficient computational power. From RGB static representations of posed FEs, approaches advanced towards dynamic representations and spontaneous expressions. In order to deal with challenges raised by large pose variations, diversity in illumination conditions and detection of subtle facial behaviour, alternative modalities like 3D and Thermal have been proposed. While most of the research focused on primary FEs and AUs, analysis of pain, fatigue, frustration or cognitive states paved the way to new applications in AFER.  

\vspace{0.3cm}
In Figure \ref{fig:history} we present a timeline of the historical evolution of AFER. In the next sections we will focus on current important trends.

\vspace{-0.2cm}\subsection{Estimating intensity of facial expressions}

While detecting FACS AUs facilitates a comprehensive analysis of the face and not only of a small subset of so called primary FEs of affect, being able to estimate the intensity of these expressions would have even greater informational value especially for the analysis of more complex facial behaviour. For example, differences in intensity and its timing can distinguish between posed and spontaneous smiles \cite{cohncohn} and between smiles perceived as polite versus those perceived as embarrassed \cite{ambadar2009all}. Moreover, intensity levels of a subset of AUs are important in determining the level of detected pain \cite{prkachin2008structure, hammal2012automatic}.

In recent years estimating intensity of facial expressions and especially of AUs has become an important trend in the community. As a consequence the Facial Expression and Recognition (FERA) challenge added a special section for intensity estimation \cite{valstar2011first, valstar2015fera}. This was recently facilitated by the publication of FE datasets that include intensity labels of spontaneous expression in RGB \cite{disfa} and 3D \cite{bp4d}.

Even though attempts in estimating FE intensity have existed before \cite{pantic2000expert}, the first seminal work was published in 2006 \cite{bartlett06}. It observed a correlation between a classifier's output margin, in this case the distance to the hyperplane of a SVM classification, and the intensity of the facial expression. Unfortunately this was only weakly observered for spontaneous FEs.

A number of studies question the validity of estimating intensity from distance to the classification hyperplane  \cite{savran2011estimation, savran2012regression, girard2014estimating}. In two works published in 2011 and 2012 Savran et al. made an excellent study of these techniques providing solutions to their main weak points \cite{savran2011estimation, savran2012regression}. They comment that such approaches are designed for AU not intensity detection and the classifier margin does not necessarily incorporate only intensity information. More recently, \cite{girard2014estimating} found that intensity-trained multiclass and regression models outperformed binary-trained classifier decision values on smile intensity estimation across multiple databases and methods for feature extraction and dimensionality reduction. 

Other works consider the possible advantage of using 3D information for intensity detection. \cite{savran2011estimation} explores a comparison between regression on SVM margins and regression on image features in RGB, 3D and their fusion. Gabor wavelets are extracted from RGB and curvatured maps from 3D captures. A feature selection step is performed from each of the modalities and from their fusion. The main assumption would be that for different AUs, either RGB or 3D representations could be more discriminative. Experiments show that 3D is not necessarily better than RGB; in fact, while 3D shows improvements on some AUs, it has performance drops on other AUs, both in the detection and intensity estimation problems. However, when 3D is fused with RGB, the overall performance increases significantly. In \cite{savran2012regression}, Savran et al. try different 3D surface representations. When evaluated comparatively, RGB representation performs better on the upper face while 3D representation performs better on the lower face and there is an overall improvement if RGB and 3D intensity estimations are fused. This might be the case because 3D sensing noise can be excessive in the eye region and 3D misses the eye texture information. On the other hand, larger deformations on the lower face make 3D more advantageous. Nevertheless, correlations on upper face are significantly higher than the lower face for both modalities. This points out to the difficulties in intensity estimation for the lower face AUs (see Figure \ref{fig:facs_aus}).

A different line of research analyzes the way geometrical and appearance representations could combine for optimizing AU intensity estimation \cite{zaker2012intensity, kaltwang2012continous}. \cite{zaker2012intensity} analyzes representations best suited for specific AUs. An assumption is made that geometrical representations perform better for AUs related to deformations (lips, eyes) and appearance features for other AUs (e.g. cheeks deformations). Testing of various descriptors is done on a small subset of specially chosen AUs but without a clear conclusion. On the other hand \cite{kaltwang2012continous} combines shape with global and local appearance features for continuous AU intensity estimation and continuous pain intensity estimation. A first conclusion is that appearance features achieve better results than shape features. Even more, the fusion between the two appearance representations, DCT and LBP, gives the best performance even though a proper alignment might improve the contribution of the shape representation as well. On the other hand this approach is static, which would fail to distinguish between eye blink and eye closure, and does not exploit the correlations between apparitions of different AUs. In order to overcome such limitations some works use probabilistic models of AUs combination likelihoods and intensity priors for improving performance \cite{sandbach2013markov, li2013unified}.

In summary, estimating facial AUs intensity followed a few distinct approaches. First, some researchers made a critical analysis about the limitations of estimating intensity from classification scores \cite{savran2011estimation, savran2012regression,girard2014estimating}. As an alternative, direct estimation from features was analyzed. Further studies on optimal representations for intensity estimation of different AUs  were published either from the points of view of geometrical vs appearance representations \cite{zaker2012intensity, kaltwang2012continous} or the fusion between RGB and 3D \cite{savran2011estimation, savran2012regression}. Finally, a third main research direction was focused on modelling the correlations between AUs appearance and intensity priors \cite{sandbach2013markov, li2013unified}. Some works are treating a limited subset of AUs while others are more extensive. All the presented approaches use predesigned representations. While the vast majority of the works are performing a global feature extraction with or without selecting features there are cases of sparse representations \cite{jeni2013continuous}. In this paper we have analyzed AU intensity estimation but significant works in estimating intensity of pain \cite{kaltwang2012continous, hammal2012automatic} or smile \cite{shimada2013fast, dhall2012group} also exist.

\vspace{-0.2cm}\subsection{Microexpressions analysis}

Microexpressions are brief FEs that people in high stake situations make when trying to conceal their feelings. They were first reported by Haggard and Issacs in 1966 \cite{haggard1966micromomentary}. Usually a microexpression lasts between 1/25 and 1/3 of a second and has low intensity. They are difficult to recognize for an untrained person. Even after extensive training, human accuracies remain low, making an automatic system highly useful. The presumed repressed character of microexpressions is valuable in detecting affective states that a person may be trying to hide. 

Microexpressions differ from other expessions not only because of their short duration but also because of their subtleness and localization. These issues have been addressed by employing specific capturing and representation techniques. Because of their short duration microexpressions may be better captured at greater than 30 fps. As with spontaneous FEs, which are shorter and less intense than exaggerated posed expressions, methods for recognizing microexpressions take into account the dynamics of the expression. For this reason, a main trend in microexpression analysis is to use appearance representations captured locally in a dynamic way \cite{shreve2009towards, shreve2011macro, pfister2011recognising}. In \cite{polikovsky2009facial} for example, the face is divided into specific regions and posed microexpressions in each region are recognized based on 3D-gradient orientation histograms extracted from sequences of frames. \cite{shreve2009towards} on the other hand use optical flow to detect strain produced on the facial surface caused by nonrigid motion. After macroexpressions have been previously detected and removed from the detection pipeline, posed microexpressions are spotted without doing classification \cite{shreve2009towards, shreve2011macro}. \cite{wu2011machine}, another method that first extracts macroexpressions before spotting microexpressions. Unlike other similar methods microexperessions are also classified into the 6 primary FEs.

A problem in the evolution of microexpression analysis has been the lack of spontaneous expression datasets. Before the publication of the CASME and the SMIC dataset in 2013, methods were usually trained with posed non-public data \cite{shreve2009towards, shreve2011macro, polikovsky2009facial}. \cite{pfister2011recognising} proposes the first microexpressions recognition system. LBP-TOP, an appearance descriptor is locally extracted from video cubes. Microexpressions detection and classification with high recognition rates are reported even at 25fps. Alternatively, existing datasets, such as BP4D, could be mined for microexpression analysis. One could identify the initial frames of discrete AUs, to mimic the duration and dynamic of microexpressions.  

In summary, microexpressions are brief, low intensity FEs believed to reflect repressed feelings. Even highly trained human experts obtain low detection rates. An automatic microexpression recognition system would be highly valuable for spotting feelings humans are trying to hide. Due to their briefness, subtleness and localization most of methods in recent years have used local, dynamic, appearance representations extracted from high frequency video for detecting and classifying posed \cite{shreve2009towards, shreve2011macro, polikovsky2009facial} and more recently spontaneous microexpressions \cite{pfister2011recognising}. 

\vspace{-0.2cm}\subsection{AFER for detecting non-primary affective states}

Most of AFER was used for predicting primary affective states of basic emotions, such as anger or happiness but FEs were also used for predicting non-primary affective states such as complex mental states \cite{el2005real}, fatigue \cite{ji2006aprobabilistic}, frustration \cite{kapoor2007automatic}, pain \cite{littlewort2007faces, ashraf09, littlewort09}, depression \cite{girard2014nonverbal, lucastowards}, mood and personality traits \cite{biel2012facetube, sanchez2013inferring}.

Approaches related to mood prediction from facial cues have pursued both descriptive (e.g., FACS) and judgmental approaches to affect. In a paper from 2009, Cohn et al. studied the difference between directly predicting depression from video by using a global geometrical representation (AAM), indirectly predicting depression from video by analyzing previously detected facial AUs and prediction depression from audio cues \cite{cohn09}. They concluded that specific AUs have higher predictive power for depression than others suggesting the advantage of using indirect representations for depression prediction. The AVEC, a challenge, is dedicated to dimensional prediction of affect (valance, arousal, dominance) and depression level prediction. The approaches dedicated to depression prediction are mainly using direct representations from video without detecting primitive FEs or AUs \cite{williamson2014vocal, sidorov2014emotion, senoussaoui2014model, jain2014depression}. They are based on local, dynamic representations of appearance (LBP-TOP or variants) for modelling continuous classification problems. Multimodality is central in such approaches either by applying early fusion \cite{senoussaoui2014model} or late fusion \cite{jain2014depression} with audio representations.

As humans rely heavily on facial cues to make judgments about others, it was assumed that personality could be inferred from FEs as well. Usually studies about personality are based on the BigFive personality trait model which is organized along five factors: openness, conscientiousness, extraversion, agreeableness, and neuroticism. While there are works on detecting personality and mood from FEs only \cite{biel2012facetube, sanchez2013inferring} the dominant approach is to use multimodality either by combining acoustic with visual cues \cite{biel2012facetube, biel2013hi} or physiological with visual cues \cite{abadi2015inference}. Visual cues can refer to eye gaze \cite{batrinca2011please, biel2012youtube}, frowning, head orientation, mouth fidgeting \cite{batrinca2011please}, primary FEs \cite{biel2012facetube, sanchez2013inferring} or characteristics of primary FEs like presence, frequency or duration \cite{biel2012facetube}. In \cite{biel2012facetube}, Biel et al. use the detection of 6 primary FEs and of smile to build various measures of expression duration or frequency. They show that using FEs is achieving better results than more basic visual activity measures like gaze activity and overall motion of the head and body; however performance is considerably worse than when estimating personality from audio and especially from prosodic cues.

In summary, in recent years, the analysis of non-primary affective states mainly focused on predicting depression. For predicting levels of depression, local, dynamic representations of appearance were usually combined with acoustic representations \cite{williamson2014vocal, sidorov2014emotion, senoussaoui2014model, jain2014depression}. Studies of FEs for predicting personality traits had mixed conclusions until now. First, FEs were proven to correlate better than visual activity with personality traits \cite{cohn09}. Practically though, while many studies have showed improvements of prediction when combined with physiological or acoustic cues, FEs remain marginal in the study of personality trait prediction \cite{biel2012facetube, batrinca2011please, biel2013hi, biel2012youtube}.

\vspace{-0.2cm}\subsection{AFER in naturalistic environments}

Until recently AFER was mostly performed in controlled environments. The publication of two important naturalistic datasets, AMFED and AFEW marked an increasing interest in naturalistic environment analysis. AFEW, \emph{Acted Facial Expressions in the Wild} dataset contains a collection of sequences from movies labelled for primitive FEs, pose, age and gender among others \cite{afew}. Additional data about context is extracted from subtitles for persons with hearing impairment. AMFED on the other hand, contains videos recording reactions to media content over the Internet. It mostly focuses on boosting research about how attitude to online media consumption can be predicted from facial reactions. Labels of AUs, primitive FEs, smiles, head movements and self reports about familiarity, liking and disposal to rewatch the content are provided. 

FEs in naturalistic environments are unposed and typically of low to moderate intensity and may have multiple apexes (peaks in intensity). Large head pose and illumination diversity are common. Face detection and alignment is highly challenging in this context, but vital for eliminating rigid motion and head pose from facial expressions. Not surprisingly, in an analysis of errors in AU detection in three-person social interactions, \cite{girard2014} found that head yaw greater than 20 degrees was a prime source of error. Pixel intensity and skin color, by contrast, were relatively benign.

While approaches to FE detection in naturalistic environments using static representations exist \cite{gehrig2013facial, dhall2011static}, dynamic representations are dominant \cite{liu2014combining, sikka2013multiple, liu2014learning, walecki15, liu2013partial, kahou2013combining}. This follows the tendency in spontaneous FE recognition in controlled environments where dynamic representations improve the ability to distinguish between subtle expressions. In \cite{liu2014learning}, spatio-temporal manifolds of low level features are modelled, \cite{sikka2013multiple} uses a maximum of a BoW (Bag of Words) pyramid over the whole sequence, \cite{kahou2013combining} captures spatio-temporal information through autoencoders  and \cite{walecki15} uses CRFs to model expression dynamics. 

Some of the approaches use predesigned representations \cite{gehrig2013facial, sikka2013multiple, dhall2011static, liu2013partial, dhall2014emotion} while recent successful approaches learn the best representation \cite{liu2014facial, kahou2013combining, liu2014learning} or combine predesigned and learned features \cite{liu2014combining}. Because of the need to detect subtle changes in the facial configuration, predesigned representations use appearance features extracted either globally or locally. Gehrig et al. in their analysis of the challenges of naturalistic environments use DCT, LBP and Gabor Filters \cite{gehrig2013facial}, Sikka et al. use dense multi-scale SIFT BoWs, LPQ-TOP, HOG, PHOG and GIST to get additional information about context \cite{sikka2013multiple}, Dhall et al. use LBP, HOG and PHOG in their baseline for the SFEW dataset (static images extracted from AFEW) \cite{dhall2011static} and LBP-TOP in their baseline for the EmotiW 2014 challenge \cite{dhall2014emotion}, and Liu et al. use convolution filters for producing mid-level features \cite{liu2014learning}.

Some representative approaches using learned representation were recently proposed \cite{liu2014facial, kahou2013combining, liu2014learning,liu2014combining}. In \cite{liu2014facial}, a BDBN framework for learning and selecting features is proposed. It is best suited for characterizing expression-related facial changes. \cite{kahou2013combining} proposes a configuration obtained by late fusing spatio-temporal activity recognition with audio cues, a dictionary of features extracted from the mouth region and a deep neural network for FEs recognition. In \cite{liu2014combining}, predesigned (HOG, SIFT) and learned (deep CNN features) representations are combined and different image set models are used to represent the video sequences on a Riemannian manifold. In the end, late fusion of classifiers based on different kernel methods (SVM, Logistic Regression, Partial Least Squares) and different modalities (audio and video) is conducted for final recognition results. Finally, \cite{walecki15} encodes dynamics with a \emph{Variable-State Latent Conditional Random Fields} (VSL-CRF) model that automatically selects the optimal latent states and their intensity for each sequence and target class. 

Most approaches presented target primitive FEs. Methods for recognizing other affective states have also been proposed, namely cognitive states like boredom, confusion, delight, concentration and frustration \cite{bosch2015automatic}, positive and negative affect from groups of people \cite{dhall2015more} or liking/not-linking of online media for predicting buying behaviour for marketing purposes \cite{mcduff2014automatic}.

In summary, large head pose rotations and illumination changes make FE recognition in naturalistic environments particularly challenging. FEs are by definition spontaneous, usually have low intensity, can have multiple apexes and can be difficult to distinguish from facial displays of speech. Even more, multiple persons can express FEs simultaneously. Because of the subtleness of facial configurations most predesigned representations are dynamically extracting the appearance \cite{gehrig2013facial, sikka2013multiple, liu2013partial, dhall2014emotion}. Recently successful methods learn representations \cite{liu2014facial, kahou2013combining, liu2014learning, liu2014combining} from sequences of frames. Most approaches target primitive FEs of affect, but others recognize cognitive states \cite{bosch2015automatic}, postive and negative affect from groups of people \cite{dhall2015more} and liking/not-linking of online media for predicting buying behaviour for marketing purposes \cite{mcduff2014automatic}.

\vspace{-0.2cm}\section{Discussion}
\label{sec:discussion}

By looking at faces humans extract information about each other, such as age, gender, race, and how others feel and think. Building automatic AFER systems would have tremendous benefits. Despite significant advances, automatic AFER still faces many challenges like large head pose variations, changing illumination contexts and the distinction between facial display of affect and facial display caused by speech. Finally, even when one manages to build systems that can robustly recognize FEs in naturalistic environments, it still remains difficult to interpret their meaning. In this paper we have focused in providing a general introduction into the broad field of AFER. We have started by discussing how affect can be inferred from FEs and its applications. An in-depth discussion about each step in a AFER pipeline followed, including a comprehensive taxonomy and many examples of techniques used on data captured with different video sensors (RGB, 3D, Thermal). Then, we have presented important recent evolutions in the estimation of FE intensities, recognition of microexpressions and non-primary affective states and analysis of FEs in naturalistic environments.


\textbf{Face localization and registration.} When extracting FE information, techniques vary according to both modality and temporality. Regardless of these approaches, a common pipeline has been presented which is followed by most methods, consisting of face detection, face registration, feature extraction and recognition itself. When combining multiple modalities, a fifth fusion step is added to the pipeline. Depending on the modality, this pipeline can vary slightly. For instance, face registration is not feasible for thermal imaging due to the dullness of the captured images, which in turn limits feature extraction to appearance-based techniques. The techniques applied to obtain the facial landmarks are different for RGB and 3D, being these feature detection and shape registration problems respectively. The pipeline may also vary for some methods, which may not require face alignment for some global feature-extraction techniques, and may perform feature extraction implicitly with recognition, as is the case of deep learning approaches.

The first two steps of the pipeline, face localization and 2D/3D registration, are common to many facial analysis techniques, such as face and gender recognition, age estimation and head pose recovery. This work introduces them briefly, referring the reader to more specific surveys for each topic \cite{zhang2010survey,wang2014facial,tam2013registration}. For face localization, two main families of methods have been found: face detection and face segmentation. Face detection is the most common approach, and is usually treated as a classification problem where a bounding box can either be a face or not. Segmentation techniques label the image at the pixel level. For face registration, 2D (RGB/thermal) and 3D approaches have been discussed. 2D approaches solve a feature detection problem where multiple facial features are to be located inside a facial region. This problem is approached either by directly fitting the geometry to the image, or by using deformable models defining a prototypical model of the face and its possible deformations. 3D approaches, on the other hand, consider a shape registration problem where a transform is to be found matching the captured shape to a model. Currently the main challenge is to improve registration algorithms to robustly deal with naturalistic environments. This is vital for dealing with large rotations, occlusions, multiple persons and, in the case of 3D registration, it could also be used for synthesising new faces for training neural networks.


\textbf{Feature extraction.} There are many different aproaches for extracting features. Predesigned descriptors are very common, although recently deep learning techniques such as CNN and DBN have been used, implicitly learning the relevant features along with the recognition model. While automatically learned techniques cannot be directly classified according to the nature of the described information, predesigned descriptors exploit either the facial appearance, geometry or a combination of both. Regardless of their nature, many methods exploit information either at a local level, centering on interest regions sometimes defined by AUs based on the FACS/FAP coding, or at a global level, using the whole facial region. These methods can describe either a single frame, or dynamic information. Usually, representing the differences between consecutive frames is done either through shape deformations or appearance variations. Other methods use 3D descriptors such as LBP-TOP for directly extracting features from sequences of frames.

While these types of feature extraction methods are common to all modalities, it has been found that thermal images are not fit to extract geometric information due to the dullness of the captured image. In the RGB case, geometric information is never extracted at the local static level. While it should be possible to do so, we hypothesise that current 2D registration techniques lack the level of precision required to extract useful information from local shape deformations. In the case of learned features, to the best of our knowledge, dynamic feature extraction has not been attempted. It is clearly possible to do so though, and it has been done for other problems.

In the case of AU intensity estimation many studies were published either from the point of view of geometrical vs appearance representations \cite{zaker2012intensity, kaltwang2012continous} or the fusion between RGB and 3D \cite{savran2011estimation, savran2012regression}. Because of the scarcity of intensity labeled data, to the best of our knowledge all approaches until now have used predesigned representations. While the vast majority of the works perform a global feature extraction with or without selecting features there are cases of sparse representations, most notably in the work of Jeni et al. \cite{jeni2013continuous}. Due to their briefness, subtleness and localization, most of the methods for detecting microexpressions use local, dynamic, appearance representations extracted from high frequency video. Detection and classification of posed \cite{shreve2009towards, shreve2011macro, polikovsky2009facial} and more recently spontaneous microexpressions \cite{pfister2011recognising} have been proposed. For predicting levels of depression, local, dynamic representations of appearance were usually combined with acoustic representations \cite{williamson2014vocal, sidorov2014emotion, senoussaoui2014model, jain2014depression}. Because of the subtleness of facial configurations in naturalistic environments most predesigned representations are dynamically extracting the appearance \cite{gehrig2013facial, sikka2013multiple, liu2013partial, dhall2014emotion}. Recently successful methods in naturalistic environments learn representations \cite{liu2014facial, kahou2013combining, liu2014learning, liu2014combining} from sequences of frames. As the amount of labelled data increases, learning the representations could be a future trend in intensity estimation. More complex representation schemes for recognizing spontaneous microexpressions and approaches combining RGB with other modalities, especially 3D, for microexpression analysis is also a direction we foresee.


\textbf{Recognition.} Recognition approaches infer emotions or mental states based on the extracted FE features. The vast majority of techniques use a multi-class classification model where a set of emotions (usually the six basic emotions defined by Ekman) or mental states are to be detected. A continuous approach is also possible. In the continuous case, emotions are represented as points in a pre-defined space, where usually each dimension corresponds to an expressive trait. This representation has advantages such as the ability to unsupervisedly define emotions and mental states, and discriminate subtle expression differences. The ease of interpretation of multi-class approaches made continuous approaches less frequent. Recognition is also divided into static and dynamic approaches, with static approaches being dominated by conventional classification and regression methods for categorical and continuous problems respectively. In the case of dynamic approaches, usually dynamic Bayesian Network techniques are used, but also others such as Conditional Random Forests and recurrent neural networks.

Many methods focus on recognizing a limited set of primary emotions (usually 6) \cite{aleksic06, kotsia07, sebe07, rifai12, liu2014learning, fang12, hayat13}. This is mainly due to a lack of more diverse datasets. Increasing the number of recognized expressions usually follows two main directions. First, expressions can be encoded based on FACS AUs \cite{tian01, littlewort04, koelstra10, walecki15} instead of directly being classified. This provides a comprehensive coding of FEs without directly making a judgement on their intentionality. Other methods exploit additional information provided by 3D facial data. Capturing depth information has important advantages over traditional RGB datasets. It is more invariant to rotation and illumination and captures more subtle changes on the face. This is useful for detecting microexpressions and facilitates recognizing a wider range of expressions, which would be more difficult with RGB alone.

In recent years, a critical analysis has been made about the limitations of estimating AUs intensity from classification scores \cite{savran2011estimation, savran2012regression, girard2014estimating} and estimation directly from features were analysed. Research suggests that using classifier scores for predicting intensity is conceptually wrong and that intensity levels should be directly learned from the ground truth \cite{girard2014estimating}. Some works treat a limited subset of AUs while other are more extensive. Usually we talk about AU intensity estimation, but significant works in estimating intensity of pain \cite{kaltwang2012continous, hammal2012automatic} or smile \cite{shimada2013fast, dhall2012group} also exist. Starting with the publication of the BU-3DFE dataset which provides four different intensity levels for every expression, advancements in recognizing primary expressions from 3D samples were made \cite{wang20063d,soyel2007facial,tang2008automatic,gong2009automatic,berretti20113d,zeng13,lemaire2013fully}. In naturalistic environments, most approaches target primitive FEs of affect. Methods for recognizing cognitive states \cite{bosch2015automatic}, positive and negative affect from groups of people \cite{dhall2015more} or liking/not-linking of online media for predicting buying behaviour for marketing purposes \cite{mcduff2014automatic} are also common. Probably a major trend in the future will be taking into account context and recognizing ever more complex FEs from multiple data sources. Additionally, a recent trend which remains to be further exploited is mapping faces to continuous emotional spaces.  


\textbf{Multimodal fusion.} Multimodality can enrich the representation space and improve emotion inference \cite{kuncheva04,russell03}, either by using different video sensors (RGB, Depth, Thermal) or by combining FEs with other sources such as body pose, audio, language or physiological information (brain signals, cardiovascular acivity etc.). Because the different modalities can be redundant, concatenating features might not be efficient. A common solution is to use fusion (see Section \ref{sec:taxonomy:fusion} for details). Four main fusion approaches have been identified: direct, early, late and sequential fusion, in most cases using conventional fusion techniques. Some more advanced late fusion techniques have been identified such as fuzzy inference systems and bayesian inference. The advantage of these methods lies on the introduction of complementary sources of information. For instance, the radiance at different facial regions, captured through thermal imaging, varies according to changes in the blood flow triggered by emotions \cite{wang2010natural,ioannou2014thermal}. Context (situation, interacting persons, place etc) can also improve emotion inference \cite{castellano13,martinez11}. \cite{stock07} shows that the recognition of FE is strongly influenced by the body posture and that this becomes more important as the FE is more ambiguous. In another study, it is shown that not only emotional arousal can be detected from visual cues but voice can also provide indications of specific emotions through acoustic properties such as pitch range, rhythm, and amplitude or duration changes \cite{fragopanagos2005emotion}. In the case of mood and personality traits prediction fusion of acoustic and visual cues has been extensively exploited. Conclusions were mixed. First, FEs were proven to correlate better than visual activity with personality traits \cite{cohn09}. Practically though, while many studies have showed improvements of prediction when combined with physiological or acoustic cues, FEs remain marginal in the study of personality trait prediction \cite{biel2012facetube, batrinca2011please, biel2013hi, biel2012youtube}. We think years to come will probably bring improvements towards integration of visual and non-visual modalities, like acoustic, language, gestures, or physiological data coming from wearable devices.



%


\vspace{-0.3cm}
\ifCLASSOPTIONcaptionsoff
  \newpage
\fi
\bibliographystyle{IEEEtran}
\bibliography{IEEEabrv,./main.bbl}

%
\vspace{-1.0cm}

\begin{IEEEbiography}[{\includegraphics[width=1in,height=1.25in,clip,keepaspectratio]{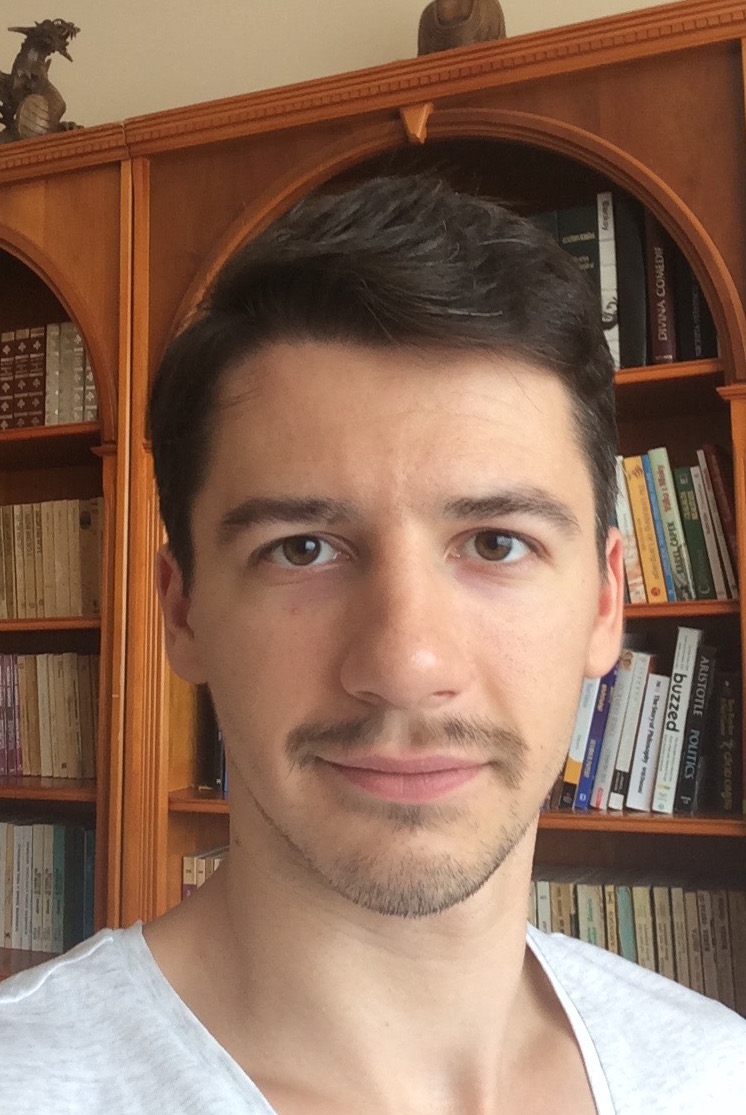}}]{Ciprian Adrian Corneanu} got his BSc in Telecommunication Engineering from T\'{e}l\'{e}com SudParis, 2011. He got his MSc in Computer Vision from Universitat Aut\'{o}noma de Barcelona. Currently he is a Ph.D. student at the Universitat de Barcelona and a fellow of the Computer Vision Center, UAB. His main research interests include face and behavior analysis, affective computing, social signal processing, human computer interaction.
\vspace{-1.0cm}\end{IEEEbiography}

\begin{IEEEbiography}[{\includegraphics[width=1in,height=1.25in,clip,keepaspectratio]{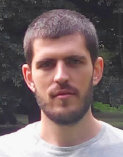}}]{Marc Oliu Sim\'{o}n}
finished his M.D in Computer Sciences and MSc in Artificial Intelligence at the Universitat Politecnica de Catalunya in 2014. Currently he is a Ph.D. student at the Universitat de Barcelona and works as a researcher at the Computer Vision Center, UAB. His main research interests include face and behaviour analysis, affective computing and neural networks.\vspace{-1.0cm}
\end{IEEEbiography}


\begin{IEEEbiography}
[{\includegraphics[width=1in,height=1.25in,clip,keepaspectratio]{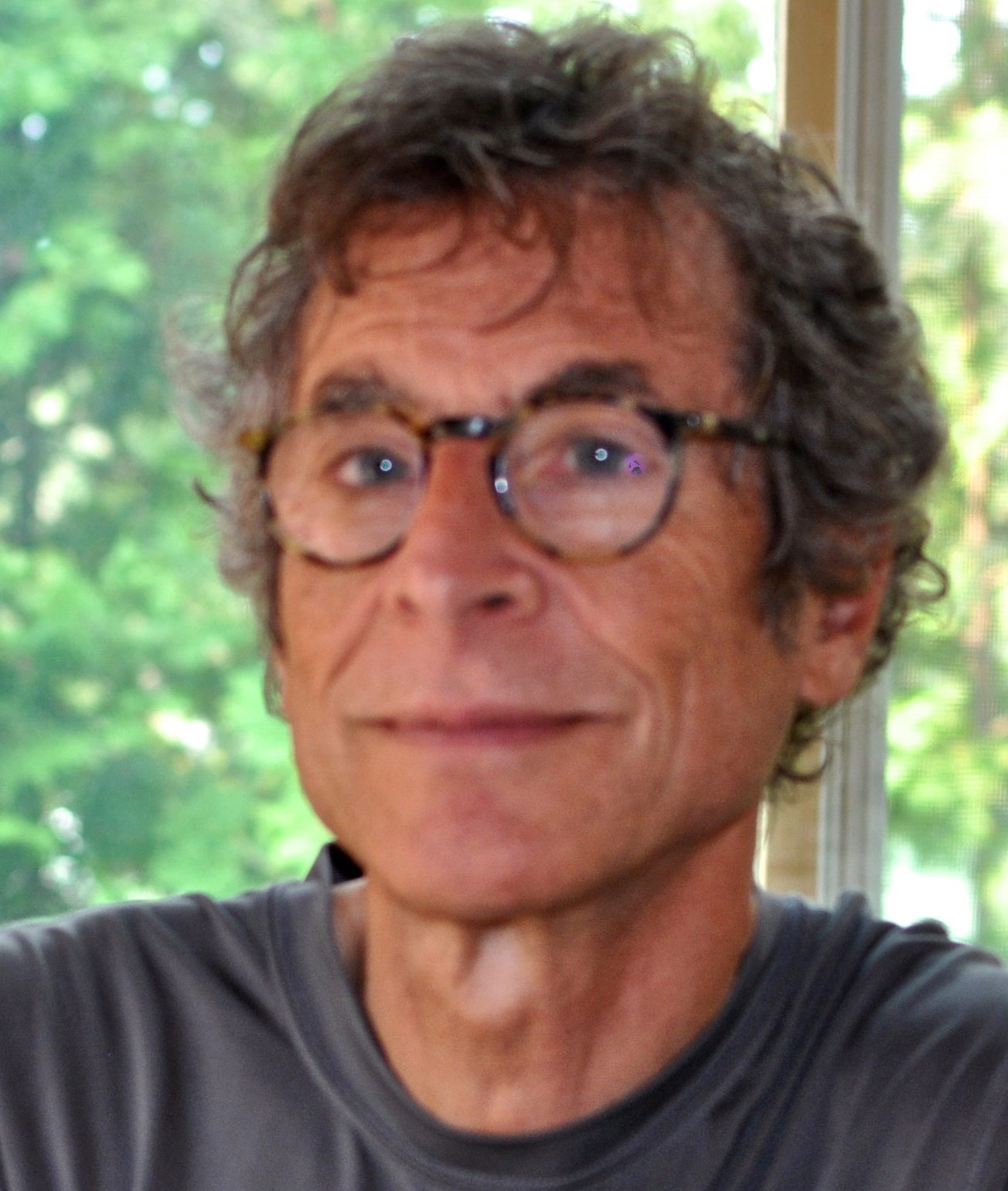}}]{Jeffrey F Cohn} is Professor of Psychology and Psychiatry at the University of Pittsburgh and Adjunct Professor of Computer Science at the Robotics Institute at CMU. He leads interdisciplinary and inter-institutional efforts to develop advanced methods of automatic analysis and synthesis of facial expression and prosody; and applies those tools to research in human emotion, social development, non-verbal communication, psychopathology, and biomedicine. His research has been supported by grants from NIH, National Science Foundation, Autism Foundation, Office of Naval Research, and Defense Advanced Research Projects Agency.\vspace{-1.0cm}
\end{IEEEbiography}

\begin{IEEEbiography}[{\includegraphics[width=1in,height=1.25in,clip,keepaspectratio]{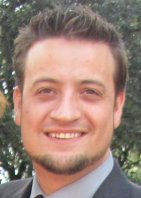}}]{Sergio Escalera Guerrero}
received his Ph.D. degree on Multiclass visual categorization systems at Computer Vision Center, UAB. He leads the HuPBA group. He is an associate professor at the Department of Applied Mathematics and Analysis, Universitat de Barcelona. He is member of the Computer Vision Center. He is director of ChaLearn Challenges in Machine Learning and vice-chair of IAPR TC-12. His research interests include, among others, statistical pattern recognition, visual object recognition, and HCI systems, with special interest in human pose recovery and behaviour analysis from multimodal data.\vspace{-1.0cm}
\end{IEEEbiography}

\vfill


\end{document}